\documentclass[sigconf]{acmart}

\usepackage{subfigure}
\usepackage{CJKutf8}
\usepackage{extarrows}
\usepackage{multirow}
\usepackage{hyperref}
\usepackage{url}
\usepackage{booktabs}

\makeatletter
\g@addto@macro\normalsize{%
  \abovedisplayskip 0pt plus2pt %minus1pt%
  \belowdisplayskip
  \abovedisplayskip
  \abovedisplayshortskip  0pt plus2pt%
  \belowdisplayshortskip  0pt plus0pt% minus1pt%
}
\makeatother

\AtBeginDocument{%
  \providecommand\BibTeX{{%
    \normalfont B\kern-0.5em{\scshape i\kern-0.25em b}\kern-0.8em\TeX}}}

\setcopyright{acmcopyright}
\copyrightyear{2018}
\acmYear{2018}
\acmDOI{XXXXXXX.XXXXXXX}

\acmConference[WSDM '24]{The 17th ACM International WSDM Conference}{March 4--March 8, 2024}{Mérida, Mexico}
\acmBooktitle{WSDM '24: The 17th ACM International WSDM Conference,
	March 4--March 8, 2024, Mérida, Mexico}
\acmPrice{15.00}
\acmISBN{978-1-4503-XXXX-X/18/06}

\begin{document}

% \title{Multi-Modal Relation Extraction via Variational Hypergraph and information bottleneck and diffusion}

\title{Text-Video Retrieval via \iffalse{}Variational\fi Multi-Modal Hypergraph Networks}

\author{Qian Li}
\affiliation{
  \institution{Baidu Inc.}
  \state{Beijing}
  \country{China}}
\email{liqian39@baidu.com}

\author{Lixin Su}
\affiliation{
  \institution{Baidu Inc.}
  \state{Beijing}
  \country{China}}
\email{sulixinict@gmail.com}

\author{Jiashu Zhao}
\affiliation{
  \institution{Wilfrid Laurier University}
  \state{Waterloo}
  \country{Canada}}
\email{jzhao@wlu.ca}

\author{Long Xia}
\affiliation{
  \institution{Baidu Inc.}
  \state{Beijing}
  \country{China}}
\email{long.phil.xia@gmail.com}

\author{Hengyi Cai}
\affiliation{
  \institution{Institute of Computing Technology, CAS}
  \state{Beijing}
  \country{China}}
\email{caihengyi@ict.ac.cn}

\author{Suqi Cheng}
\affiliation{
  \institution{Baidu Inc.}
  \state{Beijing}
  \country{China}}
\email{chengsuqi@gmail.com}

\author{Hengzhu Tang}
\affiliation{
  \institution{Baidu Inc.}
  \state{Beijing}
  \country{China}}
\email{tanghengzhu@baidu.com}

\author{Junfeng Wang}
\affiliation{
  \institution{Baidu Inc.}
  \state{Beijing}
  \country{China}}
\email{wangjunfeng@baidu.com}

\author{Dawei Yin}
\affiliation{
  \institution{Baidu Inc.}
  \state{Beijing}
  \country{China}}
\email{yindawei@acm.org}
\renewcommand{\shortauthors}{Trovato et al.}

\begin{abstract}

Text-video retrieval is a challenging task that aims to identify relevant videos given textual queries. Compared to conventional textual retrieval, the main obstacle for text-video retrieval is the semantic gap between the textual nature of queries and the visual richness of video content.
Previous works primarily focus on aligning the query and the video by finely aggregating word-frame matching signals. 
Inspired by the human cognitive process of modularly judging the relevance between text and video, the judgment needs high-order matching signal due to the consecutive and complex nature of video contents.
In this paper, we propose chunk-level text-video matching, where the query chunks are extracted to describe a specific retrieval unit, and the video chunks are segmented into distinct clips from videos.
We formulate the chunk-level matching as n-ary correlations modeling between words of the query and frames of the video and introduce a multi-modal hypergraph for n-ary correlation modeling.  
By representing textual units and video frames as nodes and using hyperedges to depict their relationships, a multi-modal hypergraph is constructed. In this way, the query and the video can be aligned in a high-order semantic space. 
In addition, to enhance the model’s generalization ability, the extracted features are fed into a variational inference component for computation, obtaining the variational representation under the Gaussian distribution.
The incorporation of hypergraphs and variational inference allows our model to capture complex, n-ary interactions among textual and visual contents. 
Experimental results demonstrate that our proposed method achieves state-of-the-art performance on the text-video retrieval task.

\end{abstract}





 
\maketitle

\section{Introduction}
\label{sec:intro}

% The number of videos has soared rapidly in recent years in the web, whereas it becomes more time-consuming and difficult to find the target video.
Text-Video Retrieval (TVR) is a multi-modal task, which aims to find the most relevant videos based on the textual query.
TVR enables humans to search videos in a simple and natural manner, hence draws numerous enthusiasm from multiple research communities \cite{DBLP:journals/ijmir/ZhuJCGL23, DBLP:conf/sigir/YakovlevPAPBNP23}.  
Different from uni-modal tasks like conventional ad-hoc retrieval, text-video retrieval operates across different modalities. Therefore, text-video retrieval is a challenging task since it requires understanding on not only the content of videos and texts, but also their inter-modal correlation.
% Text-video retrieval (TVR) is a crucial task in information retrieval, which aims to identify relevant videos based on textual queries \cite{DBLP:journals/ijmir/ZhuJCGL23, DBLP:conf/sigir/YakovlevPAPBNP23}. This approach combines multimedia content to enhance the understanding and support of various cross-modal tasks, including video recommendation systems \cite{DBLP:conf/kdd/GuptaKWWSSGJSZ20, DBLP:journals/corr/abs-2302-00569} and video question answering systems \cite{DBLP:conf/aaai/XiaoYL0JC22, DBLP:conf/aaai/CherianHMR22, DBLP:conf/acl/LyuJGF23}. The field of TVR has found applications in diverse industries, including education, entertainment, and e-commerce.

\begin{figure}[t]
    \centering
    \includegraphics[width=\linewidth]{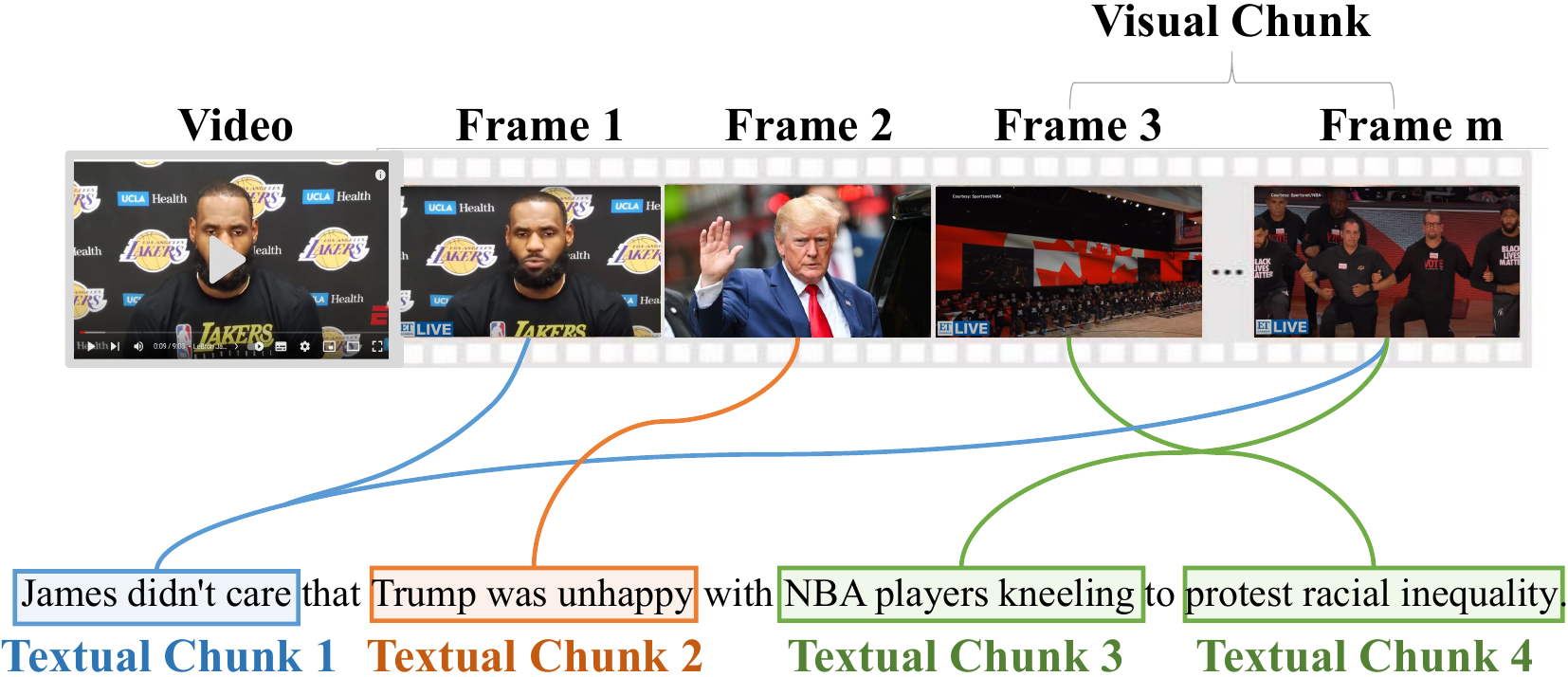}
    \caption{Text-Video Retrieval task involves identifying the intricate relations between a given text and a video. This correlation is not limited to a single word-frame correspondence but also entails complex relationships between multiple words and frames, i.e., the n-ary interactions among textual chunks and visual chunks.
    % An example of the Text-Video Retrieval task. The task is to predict the relevancy between the specific text and multiple videos.
    }
    \label{define}
\end{figure}

Cross-modality semantic representation and  alignment are at the core of the TVR task \cite{DBLP:conf/iccv/YangBG21, DBLP:conf/emnlp/LiCCGYL20}.
Existing work can mainly be divided into two categories: one focuses on cross-modal semantic representation, and the other focuses on cross-modal semantic alignment. 
Benefiting from pre-training representations, CLIP \cite{DBLP:conf/icml/RadfordKHRGASAM21} and CLIP4CLIP \cite{DBLP:journals/ijon/LuoJZCLDL22} embed the textual query and the video into shared semantic space to calculate similarity. However, these methods embed the query into a coarse representation, they lack the ability to capture fine-grained interactions. To this end, another series of works \cite{DBLP:conf/iccv/BainNVZ21, DBLP:conf/mm/MaXSYZJ22,wu2021hanet} employs attention mechanism to capture the interaction between the textual words and video frames achieving significant performance improvements.
% or their objects 
% However, existing methods primarily focus on aligning text and video in low-level semantic units, i.e. the word and the frame. 
% Although existing works have shown promising advances on the TVR task, cross-modality semantic alignment still needs to be systematically explored.
All of these methods would learn different levels of granularity for the alignment, showing promising advances in the TVR task. However, cross-modality semantic alignment still needs to be systematically explored.

% These works
% Text-Video Retrieval (TVR) poses a significant challenge in bridging the gap between textual and visual modalities \cite{DBLP:conf/iccv/YangBG21, DBLP:conf/emnlp/LiCCGYL20}. 
% Several methods have been proposed to address this task, focusing on multi-modal feature representation, which leverages models such as Multi-modal Transformer (MMT) \cite{DBLP:conf/eccv/Gabeur0AS20} and CLIP .
% Moreover, fine-grained matching and alignment approaches, such as \cite{DBLP:journals/corr/abs-2109-04290},\cite{DBLP:conf/cvpr/WangZ021},\cite{DBLP:conf/mm/MaXSYZJ22},  \cite{DBLP:conf/iccv/Liu0QCDW21}, have been explored to align the textual query and visual content in fine-grained and coarse-grained level.
% Prior art also combine multiple approaches, exemplified by Dual Encoding \cite{DBLP:conf/cvpr/DongLXJH0W19}.

% Despite recent endeavors in text-video retrieval, 
Nevertheless, identifying the most relevant videos based on a given text query faces challenging obstacles. 
For instance in Figure~\ref{define}, the text-video retrieval task demands the model to have a delicate understanding of important details such as entities \& relations (e.g., \textit{James}, \textit{Trump}, \textit{NBA players}, and \textit{kneeling}) in each semantic blocks of the text caption and the video,
i.e., representing the text or video requires the modular understanding of primitive concepts that make them.
Moreover, capturing the complex correlation between the video and text involves not only aforementioned fine-grained word-to-frame correspondences, but also n-ary interactions ({textual chunk 3 and 4} correlate with {visual chunk, i.e., frame 3 and m}, as shown in Figure~\ref{define}).
The textual semantics of the combination of the {textual chunk 3} and {textual chunk 4} conveys the event of the \textit{NBA players protest racial inequality}, and the visual chunk consists of frame 3 and frame m has the same implication.
Through understanding both the textual chunk and visual chunk, we can easily discern their relation. 
This phenomenon illustrates the vital role of multi-modal chunks in facilitating the text-video retrieval task, which remains underexplored.
% making architectures dedicated to systematic text-video modeling a necessity.

% which poses great challenges to text-video retrieval.
% which unveils an urgent need of better capturing high-order correlations between text and video.
% ultimately leading to performance degradation.

When finding the most relevant video based on the text query, although semantic features are scattered across multiple words/frames, people are capable of recognizing and clustering the key conceptual chunks presented in the text/video\cite{choi2002users}, and then judge the text-video relevance by finely aggregating the semantic relations between the recognized feature chunks.
Inspired by this observation, in this paper, we conceptualize mu\underline{\textbf{L}}ti-modal Hypergraph n\underline{\textbf{E}}tworks with v\underline{\textbf{A}}riational i\underline{\textbf{N}}ference {(LEAN)} as an approach to mimic this mechanism by constructing a multi-modal hypergraph for each text query and corresponding video, which effectively captures n-ary correlations between different modalities.
Specifically, we first construct a multi-modal hypergraph for each training instance, where textual units and video (frames in the video) are represented as a hypergraph. Next, the model automatically learns the weights of hypergraph's edges and nodes to capture the underlying associations between the words and frames. To enhance the model's generalization ability, we feed the hypergraph into a variational inference module that transforms node and hyperedge representations into Gaussian distributions.
By converting representations into Gaussian distributions, the model can better capture the underlying distribution of relationships for the retrieval task. 
The incorporation of hypergraphs and variational inference allows our model to capture complex, n-ary interactions among texts and videos and incorporate multiple type correlations, making it a promising approach for real-world applications. 
Our proposed model achieves state-of-the-art performance on TVR. Our contributions are as follows:
\begin{itemize}
    \item We technically design a novel text-video retrieval framework to capture complex and n-ary interactions among texts and videos. To the best of our knowledge, this is the first work to introduce the hypergraph networks in the text-video retrieval task.
    \item We construct a multi-modal hypergraph with three types of hyperedges for incorporating intra- and inter-modal correlations and design the multi-modal hypergraph network to capture the underlying associations between the words and frames.
     We also introduce a variational inference based graph representation learning approach to enhance the model's generalization ability.
    \item Extensive empirical results on benchmark datasets demonstrate the effectiveness of our approach, showing superior performance against state-of-the-art methods.
    
\end{itemize}

% In addition to that, several approaches have been employed, incorporating various methods, to improve performance further. For instance, Dong et al. \cite{DBLP:conf/cvpr/DongLXJH0W19} introduced a dual encoding network that enhances video features by encoding them into a highly robust dense representation and subsequently performs sequence-to-sequence matching. These notable contributions have demonstrated significant advancements in the field of text-video retrieval, exhibiting promising outcomes across different benchmark datasets.

\begin{figure*}[t]
    \centering
    \includegraphics[width=\linewidth]{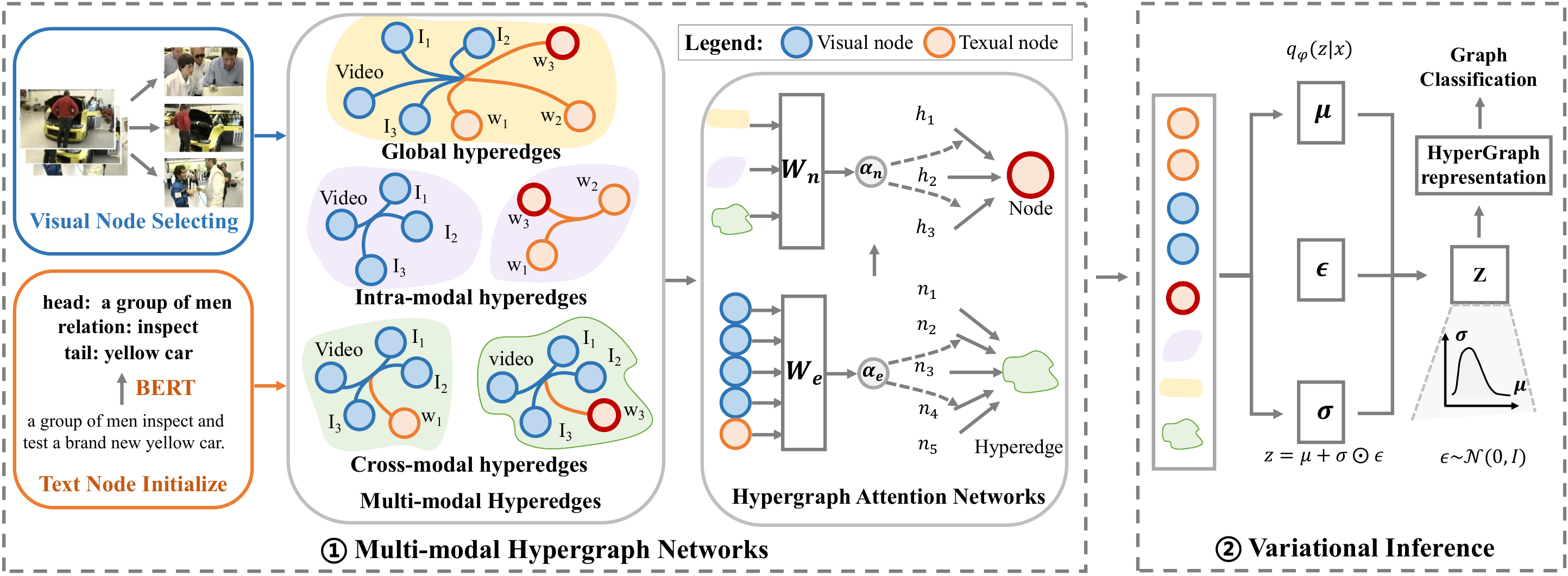}
    \caption{
    % Our framework for the text-video retrieval.
    The framework of LEAN. \textcircled{1} Multi-modal Hypergraph Networks: The hypergraph consists of video, multiple images (frames), multiple word nodes, and three types of multi-modal hyperedges for modeling chunk-level correlations. Hyperedges and nodes are updated by Hypergraph Attention Networks for n-ary interactions optimization.
    % Our framework models a text and video into a hypergraph for capturing high-order correlations by multi-modal hypergraph networks and further learns graph classification under Gaussian distribution for robust learning. 
    \textcircled{2} Variational Inference: The hypergraph representation is transferred into Gaussian distributions for better capture underlying distribution of relationships. 
    % In addition, the retrieval task is transferred to a graph classification task for predicting whether the hypergraph is valid.     
    }
    \label{Framework}
\end{figure*}

% In the context of the Text-Video Retrieval task, images inherently provide valuable information. Nonetheless, current methods employed in this domain are inadequate in addressing the issue of polysemy associated with entities and relationships. To overcome this limitation, this paper presents an innovative text-video retrieval approach, which leverages uncertain multi-modal hypergraph networks.
% The proposed method incorporates the construction of multi-modal hypergraphs for each text query and its corresponding video, facilitating the capture of high-order correlations between different modalities. By utilizing such hypergraphs, the model can effectively represent and exploit the intricate relationships present in the text and video data.
% This novel approach addresses the existing challenges faced in text-video retrieval by introducing uncertainty and hypergraphs, enabling a more comprehensive understanding of the multimodal data. Through the integration of uncertain multi-modal hypergraph networks, this method aims to enhance the performance and accuracy of text-video retrieval systems.

\section{PRELIMINARIES}

% We briefly introduce the background knowledge of Text-Video Retrieval and Hypergraph.

\paragraph{\textbf{Text-Video Retrieval (TVR)}} 

% The task of Text-Video Retrieval involves retrieving relevant video content based on a given text query or retrieving relevant text content based on a given video query. This task is a cross-modal information retrieval problem that aims to bridge the gap between different modalities, such as text and video, and enable efficient searching and browsing of multimedia content. The goal of Text-Video Retrieval is to develop algorithms and models that can effectively match and align textual and visual information to retrieve the most relevant content for a given query. This task has numerous applications in fields such as video search, surveillance, and multimedia content analysis.

TVR is a cross-modal information retrieval task that aims to retrieve relevant videos based on a given textual query or relevant text. 
The task involves matching and aligning textual and visual contents to retrieve the most relevant videos. For the textual query, the input is a word sequence $q_t$, and the output is a set of most relevant videos $V_t$, where each video $v_i \in V_t$ is associated with the text query $q_t$.
TVR is coupled with a twin task video-text retrieval, in which, given video query, the input $q_v$ denotes the visual representation, and the output is a set of most relevant texts $T_v$, where each text  $t_i \in T_v$ is associated with the video query $q_v$. 
% delete
The symbols used in this task include $q_t$ for text query, $q_v$ for video query, $V_t$ for the set of most relevant videos associated with the text query $q_t$, and $T_v$ for the set of most relevant texts associated with the video query $q_v$.

\paragraph{\textbf{Hypergraph}}
A hypergraph is a specialized form of graph featuring with hyperedges that differs from simple graphs. 
These hyperedges connect two or more nodes and are often used to represent n-ary correlations~\cite{feng2019hypergraph}. 
A hypergraph is typically defined as $\mathcal{G}=(\mathcal{X}, \mathcal{E}, \mathcal{P})$, consisting of a node set $\mathcal{X}=\left\{x_1, x_2, \ldots, x_n\right\}$, a hyperedge set $\mathcal{E}=\left\{e_1, \ldots, e_m\right\}$, and an optional diagonal weight matrix $\boldsymbol{P} \in \mathbb{R}^{m \times m}$ that represents the weight of each hyperedge. 
The hypergraph $\mathcal{G}$ can be represented by an incidence matrix $\boldsymbol{H} \in {0,1}^{n \times m}$, where each entry $\boldsymbol{H}_{i, j}$ is defined as:
% \textcolor{red}{check here}
\begin{equation}
H_{i, j}= \begin{cases}1, & \text { if } x_i \in e_j  \\ 0, & \text { otherwise }\end{cases}
\end{equation}
In this way, each hyperedge $e_j$ connects all the associated nodes $x_i$, indicating the correlations among them. 
In this paper, we adopt the concept of hypergraph to represent n-ary correlations among the textual and visual contents, providing a powerful means to establish complex correlations among semantic chunks.

% In this paper, we extend the above aggregation formulation from traditional certain hypergraph neural networks to an uncertain one, which enables us to consider the uncertainty of node or hyperedge features in a more comprehensive manner.

\section{Our Framework}

% This section introduces our proposed framework, the variational multi-modal hypergraph networks (VMHN), for the text-video retrieval task. As illustrated in Figure~\ref{Framework}, our proposed model consists of two main modules: Multi-modal Hypergraph Networks and Variational Inference. In the Multi-modal Hypergraph module, we construct a multi-modal hypergraph for each text query and corresponding video, capturing high-order correlations between different modalities. By representing text and video as nodes and using hyperedges to capture their relationships, our method effectively addresses the challenge of matching multi-modal data.
% The Variational Inference module feed the extracted features into a variational inference module for computation, obtaining the variational representation under the same distribution. This allows our model to handle situations where there is significant ambiguity or uncertainty in the relationships between textual and visual modalities, resulting in more accurate retrieval predictions. 

This section introduces our proposed framework, 
% the Multi-modal Hypergraph Networks with variational inference,
the mu\underline{\textbf{L}}ti-modal Hypergraph n\underline{\textbf{E}}tworks with v\underline{\textbf{A}}riational i\underline{\textbf{N}}ference {(LEAN)},
for the text-video retrieval task. As depicted in Figure~\ref{Framework}, LEAN consists of two main modules: Multi-modal Hypergraph Networks and Variational Inference.
In the Multi-modal Hypergraph Networks module, we construct a multi-modal hypergraph for each text query and video, capturing n-ary correlations among different modalities. 
By representing textual units and video frames as nodes and utilizing hyperedges to represent their relationships, our method effectively mitigates the challenge of matching multi-modal data. 
To achieve an optimal hypergraph structure, our model automatically learns the weights of the hypergraph, which captures the underlying associations between textual and visual modalities.
% To tackle the inherent uncertainty between textual and visual modalities, the extracted features are fed into a variational inference component for computation, obtaining the variational representation under the Gaussian distribution
The Variational Inference module induces the hypergraph representation following the Gaussian distribution, which enables our model to enhance the generalization ability between textual and visual modalities, thereby leading to more accurate retrieval predictions.

\subsection{Multi-Modal Hypergraph Networks}
Text-Video Retrieval entails extracting relationships from both textual and visual modalities.
To effectively capture intricate and high-order correlations among different modalities, we propose a Multi-modal Hypergraph Networks module. This module is designed to facilitate the understanding of high-order relations by leveraging n-ary pivotal correlations. 
More specifically, the module comprises three main components: Node Selecting and Initialization, Multi-modal Hyperedges, and Hypergraph Attention Networks.

\subsubsection{Node Selecting and Initialization}
To enhance the integration of text and video information, we selectively designate essential contents as the nodes of the hypergraph. Let $\mathcal{V}$ denote the input video and $\mathcal{S}$ denote the corresponding input textual query. 
The Multi-modal Hypergraph contains visual nodes and textual nodes.

% \textcolor{red}{VIDEO NODE.}
For visual nodes, different approaches are employed for the detection of key frames, wherein frames demonstrating notable disparities in the video content are chosen, which are selected as additional visual information. Let $\mathcal{I} = {I_{1}, I_{2}, \dots, I_{m}}$ denote the set of frames detected in video $V$. These frames are then initialized using VGG16 as follows:
\begin{equation}
x^{v_{k},0} = \mathrm{VGG16}(I_{k}), \quad \forall  k \in [1,m]
\end{equation}
where $x^{v_{k},0}$ denotes the visual feature vector for frame $I_{k}$. Furthermore, there is a global visual node of the hypergraph for aggregating all key frames, which is initialized by average pooling \cite{DBLP:journals/dsp/ZubairYW13} for compressing redundant visual information.

For textual nodes, we select triples (text chunks) as nodes for constructing the hypergraph.
Specifically, we use the Stanford coreNLP tool\footnote{\url{http://corenlp.run/}} to identify triples in each sentence. We then use BERT~\cite{DBLP:conf/naacl/DevlinCLT19} to learn contextual representations of the triples, which are used to initialize the textual nodes as follows:
\begin{equation}
 \quad x^{w_{k},0} = \mathrm{BERT}(S,w_{k}), \quad \forall k \in [1,n]
\end{equation}
where $x^{w_{k},0}$ denotes the textual feature vectors of the word $w_{k}$ in sentence $S$. $w_{k}$ denotes the $k$-th word in the selected triple in input sentence $S$, where $n$ is the number of textual nodes.

\subsubsection{Multi-modal Hyperedges}
The Multi-modal Hypergraph includes video nodes, frame nodes and textual nodes. These nodes are connected through hyperedges to establish n-ary pivotal correlations. To capture high-order correlations between text and video, we design three type hyperedges: global hyperedges, intra-modal hyperedges, and cross-modal hyperedges.

\textbf{Global hyperedges.} To infer the potential correlations and similarities that may exist between nodes, we design global hyperedges. It connects all nodes in the hypergraph, capturing global correlations among all modalities. Specifically, we define the input of global hyperedge as follows:
\begin{equation}
\mathcal{E}_{\mathrm{global}} = {(V, I_{1},  I_{2}, \dots, I_{m}, w_{1}, w_{2}, \dots, w_{n})},
\end{equation}
where each hyperedge connects $m$ visual nodes and $n$ textual nodes. 
% ~{\color{red} 1. explain i in the equation; 2. in the equation, V is the video, I is the image, neither is a VECTOR, but T is the textual feature vectors, T here should be replaced by S and E? Same questions in Intra-modal hyperedges and Cross-modal hyperedges.}

\textbf{Intra-modal hyperedges.} To deepen the understanding of the text and the visual content, we design the intra-modal hyperedges. It connects nodes within each modality, capturing correlations within each modality. Specifically, we construct two intra-modal hyperedges: one connecting the textual nodes and another connecting the video and frame nodes. The intra-modal hyperedges are defined as follows:
\begin{equation}
\mathcal{E}_{\text{text}} = {(w_{1}, w_{2}, \dots, w_{n})},
\end{equation}
\begin{equation}
\mathcal{E}_{\text{visual}} = {(V, I_{1},  I_{2}, \dots, I_{m})},
\end{equation}
where $\mathcal{E}_{\text{text}}$ denotes the intra-modal hyperedge connecting the textual nodes, and $\mathcal{E}_{\text{visual}}$ denotes the intra-modal hyperedge connecting the video and the top three frames. $w_{n}$ represents the $n$-th textual node, and $I_{m}$ denotes the $m$-th visual node.

\textbf{Cross-modal hyperedges.} To reveal the interactions among different data types, we design the cross-model hyperedges. It connects nodes across modalities, capturing correlations between different modalities. We construct $n$ types of cross-modal hyperedges.
The textual nodes connect with the video and $m$ frames. The cross-modal hyperedges are defined as follows:
\begin{equation}
\mathcal{E}_{\text{text-visual}} = {(w_{j}, V, I_{1},  I_{2}, \dots, I_{m})},
\end{equation}
where $\mathcal{E}_{\text{text-visual}}$ denotes the cross-modal hyperedge connecting the textual node $w_{j}$ with the video node $V$ and frame node $I_i$, $i\in[1, m]$.

The advantages of our proposed Multi-modal Hypergraph Construction are twofold. First, by explicitly modeling the high-order correlations among different modalities, our module can capture more complex relations that might be overlooked by existing approaches. 
Second, semantic chunks composed of multiple words and frames is able to provide useful retrieval signals.
%Second, by introducing hyperedges to connect multiple nodes simultaneously, our module can reduce the number of parameters needed to model the relationships among all nodes, which can reduce the risk of overfitting and improve the generalization performance of the model.

\subsubsection{Hypergraph Attention Networks}
To enhance the representations in a hypergraph structure, we propose a Hypergraph Attention Network, which aims to learn node representations that can effectively update both the nodes and hyperedges.
It consists of two main components: Hypergraph Encoder and Hypergraph Attention module. Hypergraph Encoder is responsible for encoding the nodes in the hypergraph, while the Hypergraph Attention module is responsible for propagating the node information through hyperedges and updating node representations.

\textbf{Hypergraph Encoder.}
The hypergraph encoder is designed to facilitate the propagation of information through hyperedges, enabling the nodes to capture the collective knowledge and features of their neighboring nodes. 
% In order to capture the inherent structure and dependencies presented in the hypergraph and effectively update node representations, we have developed the hypergraph encoder.
% This process is crucial for accurately representing the interconnections and relationships within the hypergraph.
Specifically, the equations for node updating are as follows:
\begin{equation}
\begin{aligned}
&\boldsymbol{x}_{i}^{l}=\delta\left( \sum_{j \in \mathcal{N}(i) } \frac{\boldsymbol{x}_{j}^{l-1}}{\bar{D}_{r, j }} \overline{\boldsymbol{W}}_{\mathcal{N}}^{l-1}+x_{i}^{l-1} \overline{\boldsymbol{W}}_{x}^{l-1}\right), 
\end{aligned}
\end{equation}
where $\boldsymbol{x}_i^{l-1}$ is the current representation of node $i$ in layer $l-1$, and $\overline{\boldsymbol{W}}_{\mathcal{N}}^{l-1}$ and $\overline{\boldsymbol{W}}_{x}^{l-1}$ are the learnable parameters, and
% $\mathcal{R}$ is the set of hyperedges, 
$\mathcal{N}(i)$ is the set of neighboring nodes of $i$, and $\bar{D}_{r, j}$ is a normalization factor for node $j$.
The node embeddings are updated by incorporating information from the neighboring nodes. 
By considering the collective knowledge and features of the nearby nodes, the updated node embeddings become more informative and can capture the essential characteristics of the hypergraph structure. 

To enhance understanding the role of hyperedges within the hypergraph, we update hyperedge embeddings by incorporating information from connected nodes. 
This update process enables us to capture the collective knowledge and features of interconnected nodes, thus gaining a more comprehensive understanding of the hyperedge's significance.
% The update equations for hyperedges are as follows:
We update hyperedges as follows:
\begin{equation}
\boldsymbol{e}_{j}^{w,l}=\operatorname{LeakyReLU}\left(\boldsymbol{W}_{w}^{w,l} \cdot \boldsymbol{x}_{i}^{l}\right),
\end{equation}
% \begin{equation}
% e_{j}^{v,l}=\operatorname{LeakyReLU}\left(W_{v}^{l} \cdot x_{i}^{v,l} \right),
% \end{equation}
% \begin{equation}
% e_{j}^{t,l}=\operatorname{LeakyReLU}\left(W_{t}^{l} \cdot x_{i}^{t,l} \right),
% \end{equation}
\begin{equation}
\boldsymbol{e}_{j}^{cross,l}=\operatorname{Normalization}\left(\left(\boldsymbol{x}_{i}^{t,l} \cdot(\boldsymbol{x}_{i}^{v,l})^{T}\right) \cdot \boldsymbol{x}_{i}^{l}\right),
\end{equation}
% \begin{equation}
% \begin{aligned}
% x_{i}^{l}=\left[x_{i}^{v,l} | x_{i}^{{I}_{1},l} |x_{i}^{{I}_{2},l} | x_{i}^{{I}_{3},l} |x_{i}^{T_{1},l}| x_{i}^{T_{n},l}\right],
% \end{aligned}
% \end{equation}
where $w \in [g,v,t]$, and $\boldsymbol{x}_{i}^{g,l}=\boldsymbol{x}_{i}^{l}$ is the concatenation of all node representations connected by a hyperedge, $\boldsymbol{x}_{i}^{t,l}$ is the concatenation of all textual nodes, and $\boldsymbol{x}_{i}^{v,l}$ is the concatenation of all visual nodes. 
$\boldsymbol{e}_{j}^{g,l}$, $\boldsymbol{e}_{j}^{v,l}$, $\boldsymbol{e}_{j}^{t,l}$, and $\boldsymbol{e}_{j}^{cross,l}$ are the global, intra-modal, and cross-modal hyperedges, respectively. 
$\boldsymbol{W}_{w}^{w,l}$ is the learnable parameter.
% By utilizing these within the Hypergraph Encoder, the model is able to effectively capture the information flow and dependencies within the hypergraph, resulting in enhanced node representations.
By propagating information through the hyperedges and conducting embedding updating and hyperedge aggregation steps, the hypergraph encoder effectively captures the inherent structure and dependencies within the hypergraph. 
This enables accurate updating of node representations, contributing to a more comprehensive understanding of the hypergraph and its underlying relationships.

\textbf{Hypergraph Attention.}
% To capture the underlying associations between words and frames, we. 
To compute attention weights for each hyperedge and update node representations, the hypergraph attention networks use a message-passing mechanism. Specifically, the attention weight $\alpha_{k}$ for hyperedge $k$ is computed as follows:
\begin{equation}
\alpha_k^{l} = \frac{1}{Z}\exp\left(\text{MLP}_e(\boldsymbol{x}_{i}^{l}, \boldsymbol{x}_{j}^{l})\right),
\end{equation}
where $\text{MLP}_e$ is a shared neural network that computes the similarity between nodes $i$ and $j$ connected by hyperedge $k$. $\boldsymbol{x}_{i}^{l}$ and $\boldsymbol{x}_{j}^{l}$ are the representations of the node $i$ and $j$ in the $l$-th layer, respectively, and $Z$ is a normalization factor.

% The hypergraph attention mechanism is used to update a node by aggregating information from all hyperedges connected to it. 
With the hypergraph attention mechanism, the updated node $i$ is computed by aggregating information from all hyperedges connected to node $i$.
Specifically, the node representation is computed using the following equations:
\begin{equation}
\boldsymbol{x}_{i}^{l}= \frac{1}{Z_i}\sum_{i \in\mathcal{E}_k} \alpha_{k}^l \boldsymbol{e}_{k}^{l} + \boldsymbol{x}_{i}^{l},
\end{equation}
where $\mathcal{E}_k$ is the set of hyperedges connected to node $i$, $\boldsymbol{e}_{k}^{l}$ is the hyperedge $k$ representation in the $l$-th layer, and $Z_i$ is a normalization factor. 
The weight $\alpha_{k}^l$ allows the model to focus on the most relevant nodes when updating the node representations, capturing the dependencies and structure within the hypergraph.

Grounding on this attention mechanism, hypergraph networks are derived by considering the features of related hyperedges and nodes. 
The hyperedge feature is obtained using the attention weight vector $\tilde{\boldsymbol{\alpha}}_k^l=\exp(-\gamma\boldsymbol{x}_{k}^l)$.
The hyperedges are aggregated to obtain a global representation as follows:
\begin{equation}
\begin{aligned}
&\boldsymbol{e}_{j}^l=\delta\left( \sum_{k\in\mathcal{E}_j} \frac{\boldsymbol{x}_{k}^l \odot \tilde{\boldsymbol{\alpha}}_k^l}{\tilde{\boldsymbol{D}}_{r, k }} \tilde{\boldsymbol{W}}_{r}^l\right), 
\end{aligned}
\end{equation}
where $\tilde{\boldsymbol{D}}_{r,k}$ is a normalization factor. The generated node features account for the feature of related hyperedges, while the generated hyperedge features account for the feature of the related nodes. 
The final output node feature is denoted as $\boldsymbol{x}_{k}^{l}$. 
This aggregation process considers the contributions of the connected nodes and their associated weights.

% Overall, by iteratively updating the node representations, HAN can capture higher-order correlations among the modalities and improve its ability to model complex relationships.
% The advantage of using the HAN model is that it can capture complex correlations among the different modalities and their nodes.

\subsection{Variational Inference}
After constructing textual nodes and visual nodes in a hypergraph, the feature representation of the nodes is not only influenced by its own embedding vector but also by the feature aggregation of the neighborhood. 
This process is often noisy and hinder the node representation.
High-quality node embedding is critical for the alignment of the query and the video.
By using variational inference, stochastic distribution of the latent variables is effectively inferred from the latent space rather than the observation space. 

Variational inference is a powerful method  to generate high-quality embeddings for model's generalization ability.
% Therefore, We introduce variational inference in our method.
% Variational Inference can achieves accurate and rigorous uncertainty estimation, which is crucial for neural models. 
% It can be used to quantify the model's confidence, allowing them to validate predictions on unknown input domains. 
% Due to the various correlations between textual nodes and visual nodes, they are not independently and identically distributed. 
% Therefore, it is necessary to find a suitable distribution to describe the multi-class problem of text-video retrieval, in order to estimate the model's uncertainty for enhancing the model’s generalization ability.
Specifically, we constructed a hypergraph variational autoencoder based on the structure of the graphical variational autoencoder. The specific model is as follows. First, we constructed an encoder module to solve the following form of latent variable $\boldsymbol{Z}$:
\begin{equation}
\boldsymbol{Z}=\boldsymbol{\mu}+\boldsymbol{\varepsilon} \odot \boldsymbol{\sigma}, \boldsymbol{\varepsilon} \in N(0,1)
\end{equation}
where,
\begin{equation}
\begin{gathered}
\boldsymbol{\mu}=G C N_{\mu}(\boldsymbol{X}, \overline{\boldsymbol{A}}), \\
\log \boldsymbol{\sigma}=G C N_\sigma(\boldsymbol{X}, \overline{\boldsymbol{A}}),
\end{gathered}
\end{equation}
where $\operatorname{GCN}(\boldsymbol{X}, \overline{\boldsymbol{A}})=\overline{\boldsymbol{A}} \operatorname{ReLU}\left(\overline{\boldsymbol{A}} \mathbf{X} \boldsymbol{W}_{\mathbf{0}}\right)$,
% $W_1$ is a two-layer network structure,
$\overline{\boldsymbol{A}}=D_d^{-\frac{1}{2}} \boldsymbol{H} \boldsymbol{W} D_e^{-\frac{1}{2}} H^T D_d^{-\frac{1}{2}}$ is the symmetric normalized Laplacian matrix \cite{DBLP:conf/acl/SoaresFLK19} of the hypergraph, $D_d$ is the node degree matrix of hypergraph, $D_e$ is the hyperedge degree matrix of hypergraph, $\boldsymbol{H}$ is the adjacency matrix, $\boldsymbol{W}$ is the diagonal matrix with the weight of hyperedges as diagonal elements, $G C N_\mu(\boldsymbol{X}, \overline{\boldsymbol{A}})$ and $G C N_\sigma(\boldsymbol{X}, \overline{\boldsymbol{A}})$ shares parameters of the first layer $\boldsymbol{W}_0$. By doing so, representations transfer into a latent space for high-quality embeddings.
% for high-quality embeddings, which is better for optimization.

After constructing the above model, we can further implement variational inference based on the structure we built in the hypergraph module to approximate the posterior distribution from the perspective of Gaussian distribution. After updating and learning in the hypergraph module, we input the obtained image and textual node features into the variational module, and then obtain new features through variational inference. Finally, we input these features into an MLP for final classification.
% We utilize the graph attention network methodology for learning graph representations.
% \begin{equation}
% z=\sigma\left(\sum \alpha \_i \cdot h \_i\right)
% \end{equation}
% where $h_i$ represents the representation vector of node $i$, $\alpha_i$ denotes the attention coefficient of node $i$ (related to its neighbor nodes), and $\sigma$ refers to the activation function.
The computation of graph classification probability is as follows.
\begin{equation}
\mathrm{P}(\mathrm{y} \mid \mathrm{G(t_i,v_j)})=\text { Softmax }(\text { LeakyReLU }(\mathrm{w\boldsymbol{z}})),
\end{equation}
where $\mathrm{w}$ represents the attention parameters, $\mathrm{\boldsymbol{z}}=\sigma\left(\sum \alpha \_i \cdot h \_i\right)$ denotes the representation vector of the entire hypergraph, and $\mathrm{G(t_i,v_j)}$ is the hypergraph combined by the $i$-th text and $j$-th video.

In contrast to existing methods, we construct a hypergraph by incorporating both textual and visual data, rendering the conventional loss calculation methods unsuitable. Consequently, we employ the strategy of graph classification to evaluate the model.
The text-video retrieval loss is used to measure the error between the predicted result and the ground truth result. 
The text-video retrieval loss and video-text retrieval loss are defined as follows:
\begin{equation}
\mathcal{L}_{\text {v2t}}=-\frac{1}{B} \sum_{i=1}^B C E\left(\mathrm{P}(\mathrm{y} \mid \mathrm{G(\mathcal{S}_i,\mathcal{V}_j)}), y\right),
\end{equation}
\begin{equation}
\mathcal{L}_{\text {t2v}}=-\frac{1}{B} \sum_{i=1}^B C E\left(\mathrm{P}(\mathrm{y} \mid \mathrm{G(\mathcal{S}_j,\mathcal{V}_i)}), y\right),
\end{equation}
where $CE$ is the cross entropy. $\mathcal{S}_i$ and $\mathcal{V}_i$ are the $i$-th input text and video. $y$ is the global label.
During training, given a batch of $B$ video-text pairs, the model
will generate a $B \times B$ similarity matrix. 

% https://www.overleaf.com/project/63875d77439d3eefee6369c2

\begin{table*}[!ht]
    \centering
    \renewcommand\arraystretch{1.2}
    \resizebox{\linewidth}{!}{
    \begin{tabular}{l|l|llllll|llllll}
    \toprule
        Dataset & Method &  \multicolumn{6}{c}{\textbf{Text-to-Video Retrieval}}   &  \multicolumn{6}{c}{\textbf{Video-to-Text Retrieval}} \\ \hline
        ~ & ~ & R@1↑ & R@5↑ & R@10↑ & RSUM↑ & MdR↓ & MnR↓ & R@1↑ & R@5↑ & R@10↑ & RSUM↑ & MdR↓ & MnR↓ \\ \cmidrule{2-14}
        & CE & 20.9 & 48.8 & 62.4 & 132.1 & 6.0 & 28.2 & 20.6 & 50.3 & 64.0 & 134.9 & 5.3 & 25.1 \\ \cmidrule{2-14}
        ~ & MMT & 26.6 & 57.1 & 69.6 & 153.3 & 4.0 & 24.0 & 27.0 & 57.5 & 69.7 & 154.2 & 3.7 & 21.3 \\ \cmidrule{2-14}
        ~ & SSB & 27.4 & 56.3 & 67.7 & 151.4 & 3.0 & -  & 26.6 & 55.1 & 67.5 & 149.2 & 3.0 & - \\ \cmidrule{2-14}
          & FROZEN & 31.0 & 59.5 & 70.5 & 161.0 & 3.0 &  -  &  -  &  -  & - &  -  &  - & - \\ \cmidrule{2-14}
        MSR-VTT & CLIP4Clip-meanP & 43.1 & 70.4 & 80.8 & 194.3 & 2.0 & 16.2 & 43.1 & 70.5 & 81.2 & 194.8 & 2.0 & 12.4 \\\cmidrule{2-14}
        ~ & CLIP4Clip-seqTransf & 44.5 & 71.4 & 81.6 & 197.5 & 2.0 & 15.3 & 42.7 & 70.9 & 80.6 & 194.2 & 2.0 & 11.6 \\ \cmidrule{2-14}
        ~ & CLIP2Video & 45.6 & 72.6 & 81.7 & 199.9 & 2.0 & 14.6 & 43.5 & 72.3 & 82.1 & 197.9 & 2.0 & 10.2 \\ \cmidrule{2-14}
        ~ & X-CLIP (ViT-B/16)  & \underline{49.3} & \underline{75.8} & \underline{84.8} & - & 2.0 & 12.2 & \underline{48.9} & \underline{76.8} & 84.5 & - & 2.0 & \underline{8.1} \\\cmidrule{2-14}
        ~ & MIL-NCE & 47.2 & 73.0 & 82.8 & \underline{203.0} & 2.0 & 13.9 & 46.3 & 74.1 & \underline{84.8} & \underline{205.2} & 2.0 & 8.8  \\\cmidrule{2-14}
        ~ & DiCoSA & 47.5 & 74.7 & 83.8 & - & 2.0 & 13.2 & 46.7 & 75.2 & 84.3 &  -  & 2.0 & 8.9 \\
        % ~ & DiffusionRet & 49.0 & 75.2 & 82.7 & \textbf{206.9} & 2.0 & \underline{12.1} & 47.7 & 73.8 & 84.5 & \underline{206.0} & 2.0 & 8.8 \\
        \hline
        ~ & \textbf{LEAN (Ours)} & \textbf{50.6} & \textbf{77.1} & \textbf{85.2} & \textbf{206.3} & 2.0 & \textbf{11.4} & \textbf{49.1} & \textbf{78.2} & \textbf{86.7} & \textbf{207.9} & 2.0 & \textbf{7.3} \\

\midrule
\midrule

        Dataset & Method &  \multicolumn{6}{c}{\textbf{Text-to-Video Retrieval}}   &  \multicolumn{6}{c}{\textbf{Video-to-Text Retrieval}} \\ \hline
        ~ & ~ & R@1↑ & R@5↑ & R@10↑ & RSUM↑ & MdR↓ & MnR↓ & R@1↑ & R@5↑ & R@10↑ & RSUM↑ & MdR↓ & MnR↓ \\ \hline
         & CE & 19.8 & 49.0 & 63.8 & 132.6 & 6.0 & 23.1 &  - & - &  - & - &  - & - \\ \cmidrule{2-14}
        ~ & SSB & 28.4 & 60.0 & 72.9 & 161.3 & 4.0 &  - & - &  - & - &  - & - &  - \\ \cmidrule{2-14}
        ~ & FROZEN & 33.7 & 64.7 & 76.3 & 174.7 & 3.0 &  - & - &  - & - &  - & - &  - \\ \cmidrule{2-14}
        MSVD & CLIP4Clip-meanP & 46.2 & 76.1 & 84.6 & 206.9 & 2.0 & 10.0 & 56.6 & 79.7 & 84.3 & 220.6 & 1.0 & 7.6 \\ \cmidrule{2-14}
        ~ & CLIP2Video & 47.0 & 76.8 & 85.9 & 209.7 & 2.0 & 9.6 & 58.7 & 85.6 & 91.6 & 235.9 & 1.0 & 4.3 \\ \cmidrule{2-14}
        ~ & X-CLIP (ViT-B/16)  & \underline{50.4} & \underline{80.6} & - & - & - & \underline{8.4} & 66.8 & \underline{90.4} & - & - & - & \underline{4.2} \\ \cmidrule{2-14}
        ~ & MIL-NCE & 47.5 & 78.0 & 86.6 & \underline{212.1} & 2.0 & 9.3 & \underline{70.2} & 88.1 & \underline{92.7} & \underline{251.0} & 1.0 & 6.0 \\ \cmidrule{2-14}
        
         ~ & DiCoSA & 47.4  & 76.8 & 86.0 & - & 2.0 & 9.1 & - & - & - & - & - & - \\
        % ~ & DiffusionRet & 46.6 & 75.9 & 84.1 & 206.6 & 2.0 & 15.7 & - & - & - & - & - & - \\
        \hline
        ~ & \textbf{LEAN (Ours)} & \textbf{52.1} & \textbf{81.9} & \textbf{87.0} & \textbf{213.6} & 2.0 & \textbf{7.4} & \textbf{71.3} & \textbf{91.7} & \textbf{93.8} & \textbf{253.2} & 1.0 & \textbf{4.1} \\ \hline
        
    \end{tabular}
    }
    \caption{Main experiments. The best results are highlighted in bold, ``–'' means results are not available, and the underlined values are the second-best result.
    % ``$\uparrow$'' means the increase compared to the underlined values. 
% "$\downarrow$" means the average decrease of all four metrics compare to the ProG.
}
\label{main}

\end{table*}

In our hypergraph networks, except for the retrieval loss,
 we also introduce the variational loss:
\begin{equation}
\mathcal{L}_{\text {v}}=  \sum_{i=1}^m\left(1+\log \left(\sigma_{x_{i}}^2\right)-\mu_{x_{i}}^2-\sigma_{x_{i}}^2\right),
\end{equation}
% where $\alpha$ represents a hyperparameter weight, $CE$ denotes the cross-entropy between $X$ and $\bar{X}$ indicating the accuracy of classification, and 
The $\mathcal{L}_{\text {v}}$ represents the Kullback-Leibler loss used to measure the difference between the distributions before and after variation.

The overall loss function of the LEAN is a weighted sum of the three loss terms:
\begin{equation}
\mathcal{L}=\lambda_{\text{v2t}} \mathcal{L}_{\text{v2t}}+\lambda_{\text{t2v}} \mathcal{L}_{\text{t2v}}+\lambda_{\text{v}} \mathcal{L}_{\text{v}},
\end{equation}
where $\lambda_{\text{v2t}}$, $\lambda_{\text{t2v}}$ and $\lambda_{\text{v}}$ are hyperparameters that control the relative importance of each loss term.

\section{Experiment}
% This section conducts experiments to evaluate the performance of our proposed model.

% \begin{table}[bp]
% \centering
% \resizebox{\linewidth}{!}{
% \begin{tabular}{l|rrrrrr}
% \toprule
% \textbf{Dataset} &  \#Entity &  \#Rel. & \#Attr. & \#Rel. T. & \#Attr. T. & \#Images \\ 
% \midrule
%  \textbf{MNRE} & 14,951 & 1,345  & 116 & 592,213  &29,395 &13,444 \\ 
% \bottomrule
% \end{tabular}
% }
% \caption{Statistics for the datasets. (Rel.: Relation, Attr.: Attribute, Rel. T.: Relational Triple, Attr. T.: Attributes triple.)}
% \label{dataset}
% \end{table}

\subsection{Dataset and Evaluation Metric}

We conducted experiments on two Text-Video Retrieval datasets: MSR-VTT \cite{xu2016msr} and MSVD \cite{chen2011collecting}.
MSR-VTT (Microsoft Research Video to Text) is a large-scale dataset for open-domain video captioning, which consists of 10,000 video clips from 20 categories.
%and each video clip is annotated with 20 English sentences by Amazon Mechanical Turks. 
The objective is to retrieve video segments that match a given textual description. 
The standard splits use 6,513 clips for training, 497 clips for validation, and 2,990 clips for testing. 
MSVD consists of 1,970 videos with varying durations ranging from 1 to 62 seconds. Each video is associated with 40 English captions. In this dataset, 1,200 videos were utilized for training, while 100 videos were allocated for validation, and 670 videos were set aside for testing purposes.
% LSMDC, a dataset co-released by Johns Hopkins University and FAIR (Facebook AI Research), consists of 118,081 video segments accompanied by bilingual subtitles in English and French. The task of this dataset is to retrieve video segments that match a given textual description or bilingual subtitles.
We reported the official evaluation metrics, including R@1, R@5, R@10, RSUM, MdR, and MnR, for assessing the performance of the models in terms of retrieval accuracy and ranking evaluation, which are widely used in previous retrieval works\cite{DBLP:conf/mm/MaXSYZJ22,guo2020deep}.

% We follow the standard retrieval task and report Recall at rank K (R@K), median rank (MdR) and mean rank (MnR) as metric, where the higher R@K, and lower median rank and mean rank indicates better performance.

\subsection{Comparision Methods}

We compare our method with eight text-video retrieval models, which focus on cross-modality semantic representation and alignment among the visual contents and the textual words: 
\begin{itemize}
\item  CE \cite{DBLP:conf/bmvc/LiuANZ19} condenses high-dimensional information from videos into a compact representation using free-form text queries.
\item  MMT \cite{DBLP:conf/eccv/Gabeur0AS20} aggregates sequences of multi-modal features and models the temporal information.
% It utilizes pre-trained semantic embeddings that encompass general features like motion, appearance, and scene features from visual content.
\item  SSB \cite{DBLP:conf/iclr/Patrick0AMHHV21} uses a generative model to group related samples by reconstructing each sample's caption.
% with a support set of visual representations. 
% This approach ensures that the representations are not overly specific to individual samples, making them reusable across the dataset. 
% By explicitly encoding shared semantics between samples, SSB avoids the limitations of noise contrastive learning. 
\item FROZEN \cite{DBLP:conf/iccv/BainNVZ21} is an end-to-end trainable model that extends the ViT and Timesformer architectures.
% for leveraging large-scale image and video captioning datasets. 
% It incorporates attention mechanisms in both spatial and temporal dimensions. 
% The model's flexibility enables independent or combined training on image and video text datasets.
\item  CLIP4Clip \cite{DBLP:journals/ijon/LuoJZCLDL22} applies transfer learning from the image-to-text pre-trained CLIP model in an end-to-end manner. 
\item  CLIP2Video \cite{DBLP:journals/corr/abs-2106-11097} learns image-text interactions and improves temporal relations between video frames and video-text.
\item  X-CLIP \cite{DBLP:conf/mm/MaXSYZJ22} introduces a multi-grained contrastive model and the Attention Over Similarity Matrix module to focus on contrasting essential frames and words. 
\item  MIL-NCE \cite{DBLP:conf/sigir/LiuCHCWPW22} presents an image animation strategy that facilitates the conversion of commonly used image-language corpora into synthesized video-text data for pretraining. 
\item  DiCoSA \cite{DBLP:journals/corr/abs-2305-12218} designs Set-to-set Alignment to simulate the conceptualization and utilizes the adaptive pooling mechanism to merge semantic concepts for partial matching.
% \item DiffusionRet \cite{DBLP:journals/corr/abs-2303-09867} utilizes a generative manner for the TVR and the 
% joint probability to model the correlations between the texts and the videos.
\end{itemize}

\subsection{Implementation Details}

We used PyTorch\footnote{https://pytorch.org/} as a deep learning framework to develop the TVR. All experiments were conducted on a server with one GPU (Tesla V100).
The BERT version is bert-base-uncased in huggingface\footnote{https://github.com/huggingface/transformers} for text initialization and the dimension was set to 768. The VGG version is VGG16\footnote{https://github.com/machrisaa/tensorflow-vgg} for image initialization. The dimension of visual images features was set to 4096. The AdamW optimizer was chosen with a weight decay of 0.01. 
% The number of images extracted from the image was set to 10, and 
The batch size is set to 16.
We performed a search for the learning rate in the range of 1e-5 to 6e-5, and ultimately settled on 3e-5. Additionally, we conducted a search for the number of iterations for the hypergraph update ranging from 1 to 10, and determined the optimal value to be 2.
The training was carried out for 30 epochs, with evaluation performed after the 10-th epoch. The model that performed the best on the validation set was selected and evaluated on the test set.
To ensure fairness, all baseline models use the same data set partitioning.

\begin{table*}[t]
\centering
\renewcommand\arraystretch{1.2}
\resizebox{\linewidth}{!}{
\begin{tabular}{l|cccccc|cccccc}
\toprule
% \textbf{Tasks} & \multicolumn{4}{|c}{\textbf{Our dataset} } \\ 
 \textbf{Variants}  & R@1↑ & R@5↑ & R@10↑ & RSUM↑ & MdR↓ & MnR↓  & R@1↑ & R@5↑ & R@10↑ & RSUM↑ & MdR↓ & MnR↓ \\
\midrule
 \textbf{$\text{LEAN (Ours)}$}  & \textbf{50.6} & \textbf{77.1} & \textbf{85.2} & \textbf{206.3} & 2 & \textbf{11.4} & \textbf{49.1} & \textbf{78.2} & \textbf{86.7} & \textbf{207.9} & 2 & \textbf{7.3}  \\ \hline
 % {  }{  } {  }{  }\textbf{  w/o Uncertain Encoder}   & 48.6 & 74.7 & 85.2 & 205.5 & 2 & 11.4 & 49.1 & 75.6 & 83.9 & 207 & 2 & 7.3  \\  
  {  }{  } {  }{  }\textbf{  w/o Hypergraph Network}  & 48.4 & 75.3 & 83.6 & 205.1 & 4 & 10.2 & 48.0 & 76.9 & 84.6 & 207.3 & 3 & 6.8 \\ 
   % {  }{  } {  }{  }\textbf{ w/o Uncertain Hypergraph Network}  & 48.6 & 74.7 & 85.2 & 205.5 & 2 & 11.4 & 49.1 & 75.6 & 83.9 & 207 & 2 & 7.3 \\  
    {  }{  } {  }{  }\textbf{  repl. HGNN}   & 48.5 & 75.8 & 84.1 & 205.2 & 3 & 10.7 & 48.5 & 76.1 & 84.0 & 207.1 & 3 & 6.5 \\ 
 {  }{  } {  }{  }\textbf{  repl. Uncertain GCN}   & 49.2 & 76.3 & 84.4 & 204.9 & 3 & 10.5 & 47.8 & 75.9 & 83.5 & 206.8 & 3 & 6.9 \\ 
 {  }{  } {  }{  }\textbf{  repl. GCN}   &  49.0 & 76.2 & 84.6 & 205.8 & 2 & 10.6 & 48.7 & 77.3 & 85.2 & 206.9 & 2 & 7.1  \\ 
  {  }{  } {  }{  }\textbf{  repl. GAT}  &  49.2 & 76.1 & 84.3 & 205.9 & 2 & 10.8 & 48.3 & 77.7 & 85.6 & 206.7 & 2 & 7.0  \\ 
  {  }{  } {  }{  }\textbf{  w/o Uncertain Loss}    &  49.7 & 76.9 & 84.8 & 206.1 & 2 & 11.2 & 48.9 & 78.0 & 86.3 & 206.3 & 2 & 7.2  \\ \hline
   {  }{  } {  }{  }\textbf{ w/o Global hyperedges}   & 49.7 & 76.6 & 84.7 & 206.0 & 3 & 11.0 & 48.8 & 77.9 & 86.3 & 207.5 & 3 & 6.9\\  
  {  }{  } {  }{  }\textbf{ w/o Textual hyperedges}  & 49.9 & 76.8 & 84.8 & 205.7 & 2 & 11.3 & 48.5 & 78.0 & 86.5 & 207.8 & 3 & 6.7 \\ 
   {  }{  } {  }{  }\textbf{ w/o Visual hyperedges}  & 50.2 & 76.7 & 84.5 & 205.6 & 2 & 10.9 & 48.7 & 77.7 & 86.0 & 207.2 & 3 & 7.0  \\  
 {  }{  } {  }{  }\textbf{ w/o Cross-modal hyperedges}   & 50.1 & 76.6 & 84.9 & 205.8 & 2 & 11.1 & 48.9 & 77.6 & 86.2 & 207.5 & 3 & 7.2   \\ 
  % {  }{  } {  }{  }\textbf{ w/o tail-image hypergraph}   & 48.6 & 74.7 & 85.2 & 205.5 & 2 & 11.4 & 49.1 & 75.6 & 83.9 & 207 & 2 & 7.3   \\ 
\bottomrule
\end{tabular}
}
\caption{Variant experiments. “w/o” means removing corresponding module from the complete model. “repl.” means replacing corresponding module with the other module. }
\label{Ablation}
\end{table*}

\subsection{Main Results}
To verify the effectiveness of our model, we report overall average results in Table~\ref{main}. From the table, we can observe that: 1) Our model outperforms all the baselines in all metrics. Especially, our model improves at least 1.3\% on R@1 for the TVR and 1.5\% on RSUM, respectively. It demonstrates our model constructing a multi-modal hypergraph for each text query and video, capturing high-order correlations between different modalities. 2) Compared to fine-grained matching and alignment TVR models, our model achieves better performance, capturing the underlying associations between textual and visual modalities. 3) Compared to multi-modal transformer-based TVR models, our model performs better, learning an optimal hypergraph structure and obtaining the variational representation under the same distribution. 4) Our model also outperforms graph neural networks for TVR, indicating that our hypergraph learning method can capture more high-order correlations and generalization modeling, leading to better performance. All the observations demonstrate that our model is effective in leveraging multi-modal information and capturing complex correlations for the TVR task. It is also able to outperform various baselines and achieve state-of-the-art performance on the task.

\begin{figure}[t]
    \centering
    \includegraphics[width=\linewidth]{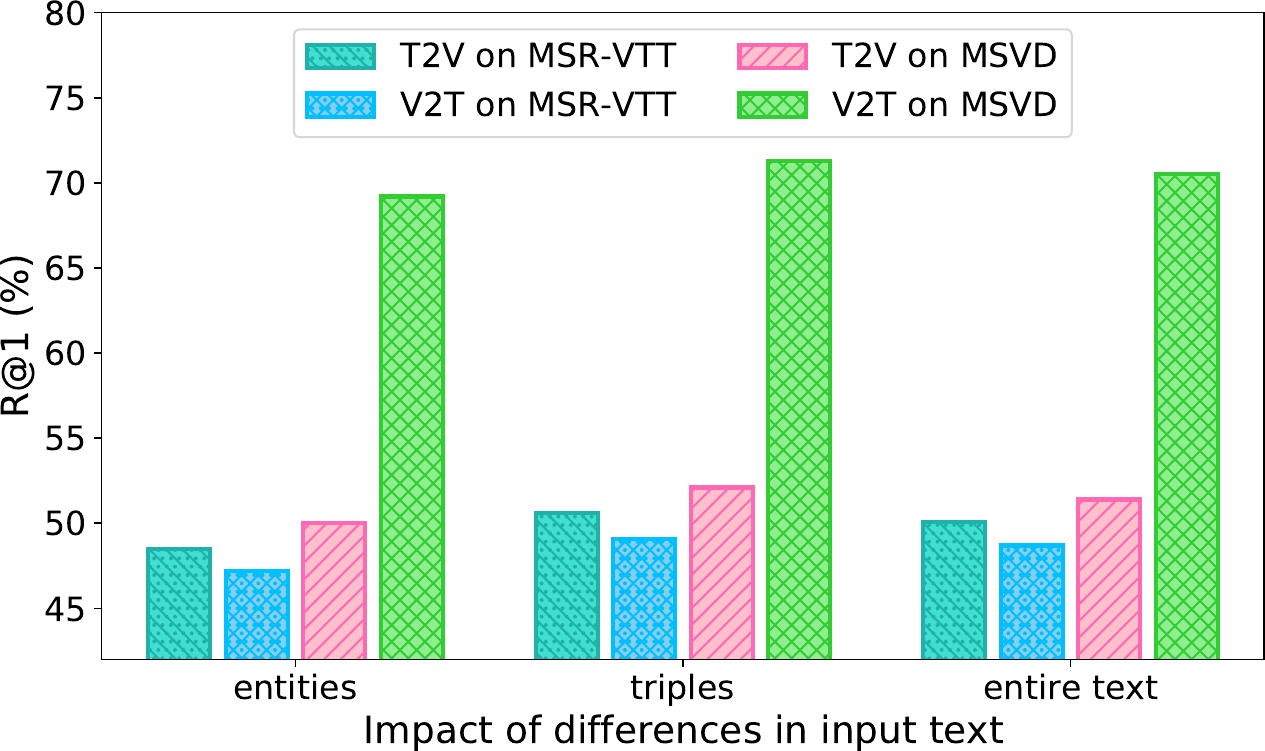}
    \caption{Impact of differences with different input text. }
    \label{Efficiency}
\end{figure}

\subsection{Discussion for Model Variants}
To investigate the effectiveness of each module in our model, we conduct variant experiments, showcasing the results in Table~\ref{Ablation}. From the table, we can observe that: 1) The impact of the multi-modal hypergraph tends to be more significant than other modules. We believe this is because the multi-modal hypergraph facilitates high-order semantic understanding by exploiting n-ary pivotal correlations, thus providing more clues for the TVR task. 2) The replacement of variational inference with specific representations resulted in a decline in performance. This indicates that our full model can better capture the underlying distribution of relationships and generate more accurate predictions. 3) By removing any one of the hypergraphs, the performance basically decreased. It demonstrates that each hypergraph is helpful for retrieval and captures different high-order semantics. 4) By replacing the Hypergraph Network with other GNN models (such as GCN) resulted in decreased performance, suggesting that n-ary correlations between different modalities and hypergraph is useful for the TVR task.
All the observations demonstrate the effectiveness of each component in our model and how they contribute to its superior performance.
% in the TVR task.

\begin{figure}[t]
    \centering
    \includegraphics[width=\linewidth]{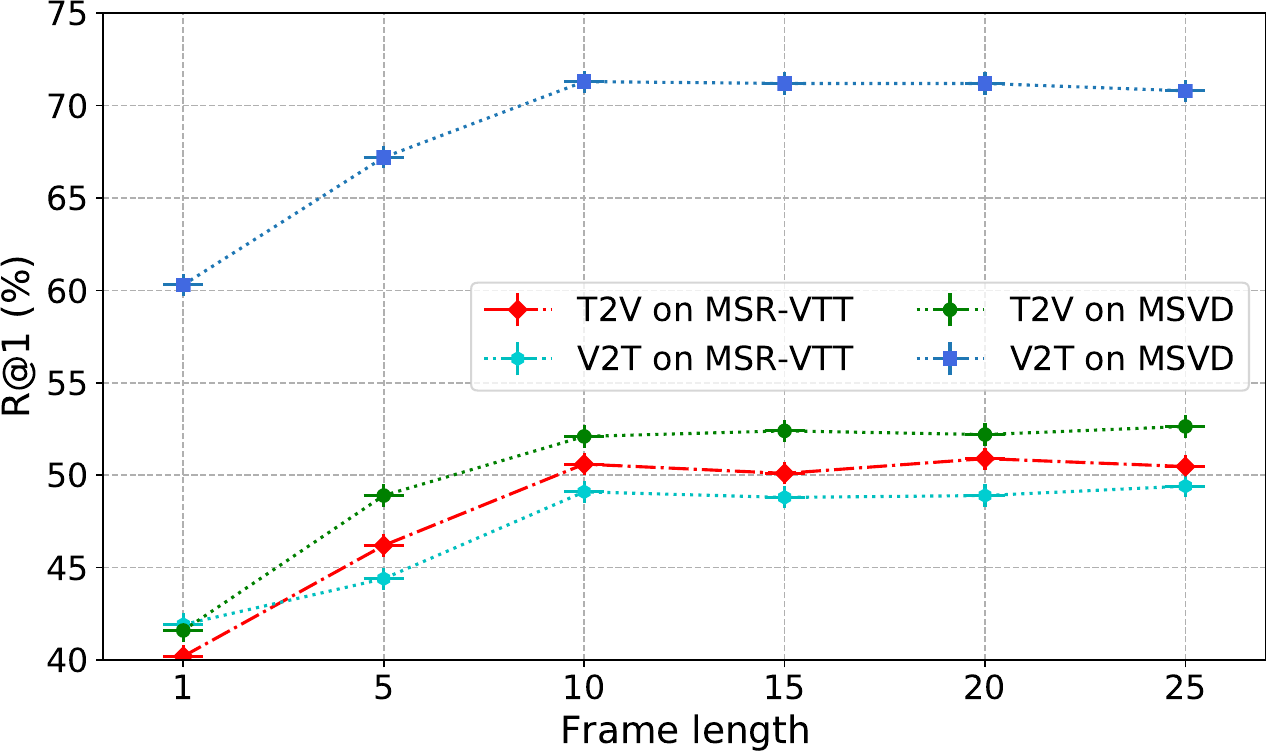}
    \caption{Impact of differences in frame length. }
    \label{Proportion}
\end{figure}

\subsection{Discussion for Hypergraph Construction}

For the input text, we conducted a comparison of the model's performance when selecting different chunks form the query as the nodes, e.g., entities, triples, and the entire text, as depicted in Figure \ref{Efficiency}. 
From the figure, we can observe that: (1) The model achieved the highest performance when event triples were chosen as the textual nodes in the hypergraph. This finding indicates that triple information is particularly valuable for the retrieval task. (2) The performance of the model when selecting the entire text was comparable to that of when triples were chosen. This implies that triples serve as the essential information for enhancing retrieval, and only utilizing event triples as nodes in the hypergraph leads to a lower computational complexity compared to using the entire text. Hence, we have selected event triples as the textual nodes in the hypergraph.

For the input video, we investigated the performance of the model by examining its performance with different numbers of frames, as shown in the Figure \ref{Proportion}. From the figure, we can observe that: (1) The performance of the model improves with an increase in the number of frames. This suggests that more effective key frames are beneficial for retrieval, and the model is capable of capturing more correlated information between the text and video through additional frames. (2) The model's performance stabilizes when the number of frames reaches 10. This indicates that within a certain range of frame numbers, the model is already able to effectively capture the correlated information between the text and video. Further increasing the number of frames does not significantly enhance the model's performance, possibly because the additional frames do not provide more useful information or the model has reached its performance limit.

\subsection{Discussion for Hyperpamaters}

\begin{figure}[t]
    \centering
    \includegraphics[width=\linewidth]{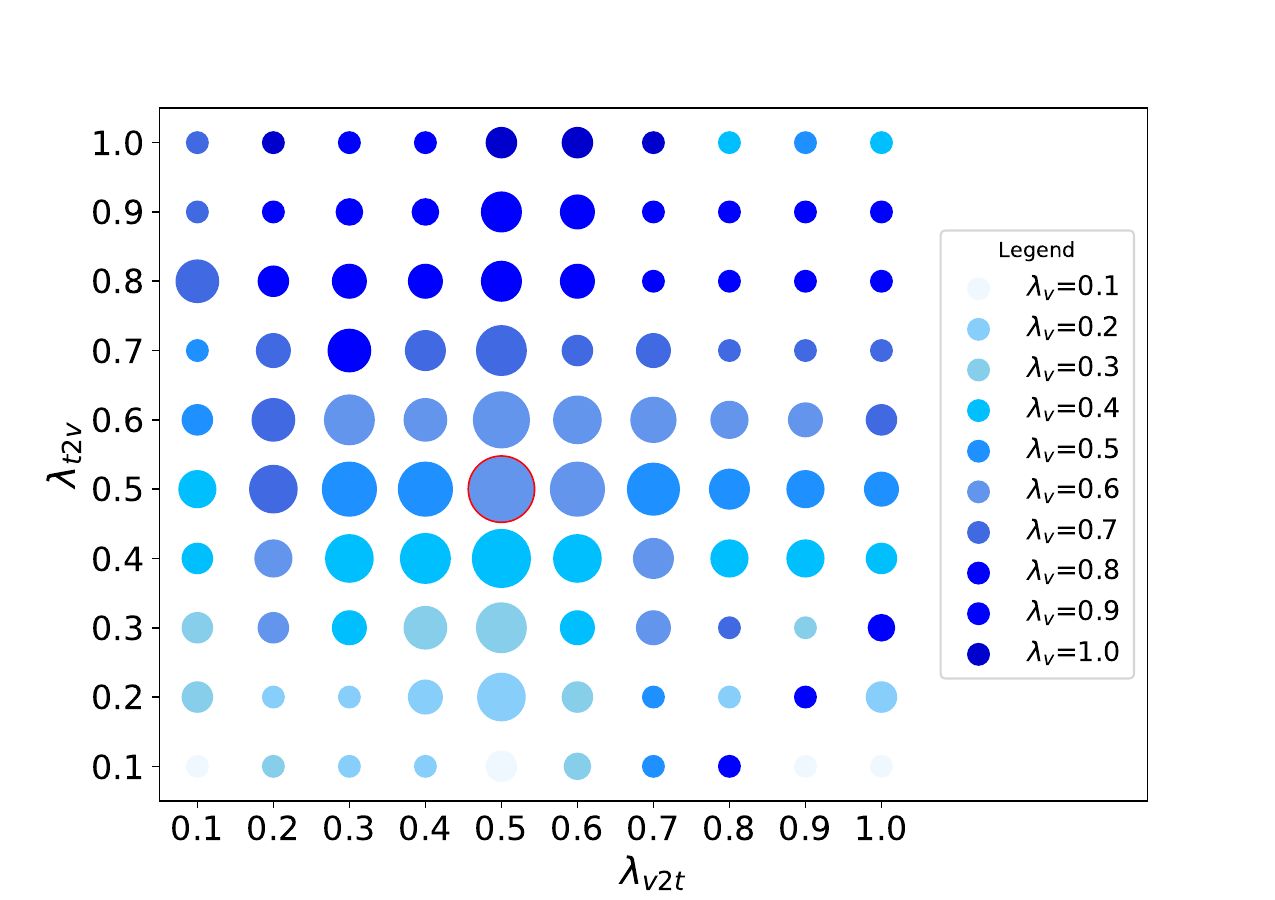}
    \caption{The performance of different hyperpamaters of $\lambda_{\text{v2t}}$, $\lambda_{\text{t2v}}$ and $\lambda_{\text{v}}$. The larger the circle, the higher the performance.
    }
    \label{caseStudy}
\end{figure}

To investigate the impact of the three loss functions on the performance of our model, we conducted experiments as shown in the Figure \ref{caseStudy}. The $\lambda_{\text{v}}$ is the best performance when the values of $\lambda_{\text{v2t}}$ and $\lambda_{\text{t2v}}$ are fixed. The size of the circles in the figure represents the quality of the model results, with larger circles indicating better performance.  The experimental results demonstrate that the model achieves optimal performance when the hyperparameters are set to 0.5, 0.5, and 0.6, respectively. Additionally, significant performance improvement is observed when the differences between the three hyperparameter values are relatively small and the coefficient of the KL loss is large. This highlights the importance of the KL loss in enhancing the model's performance.
This may be attributed to the ability of the KL loss to effectively capture the distributional differences between the predicted and target values, leading to more accurate modeling of the underlying data patterns. 

\section{Related Work}

Text-Video Retrieval (TVR) aims to retrieve relevant videos based on textual queries \cite{DBLP:journals/ijmir/ZhuJCGL23, DBLP:journals/ijon/LuoJZCLDL22, DBLP:journals/corr/abs-2305-12218,han2021visual,lei2021less,wang2022hybrid}. In recent years, numerous methods have been proposed to tackle this problem. One prominent direction in TVR is the extraction and encoding of multi-modal features from videos, including visual, audio, and textual information. This approach has been extensively studied by researchers. For instance, Gabeur et al. \cite{DBLP:conf/eccv/Gabeur0AS20} introduced a Multi-modal Transformer (MMT) model specifically designed for video retrieval by incorporating multiple modalities. Moreover, the utilization of large-scale pre-trained models has gained attention in video-text retrieval tasks. CLIP (Contrastive Language-Image Pretraining) \cite{DBLP:conf/icml/RadfordKHRGASAM21} learns joint representations of images and texts, has been adopted by Fang et al. \cite{DBLP:journals/corr/abs-2106-11097} and Luo et al. \cite{DBLP:journals/ijon/LuoJZCLDL22} in their end-to-end models for feature extraction and temporal alignment.
These advancements have paved the way for more accurate and robust video retrieval from textual queries. The integration of multi-modal features and the utilization of pre-trained models have demonstrated significant improvements, pushing the boundaries of state-of-the-art TVR systems.

In addition to the aforementioned approaches, another effective strategy in TVR is to perform fine-grained matching and alignment between video and text \cite{liu2022featinter,dong2022reading,guo2021ssan,croitoru2021teachtext}. This approach aims to capture detailed correspondences between the two modalities.
Cheng et al. \cite{DBLP:journals/corr/abs-2109-04290} proposed the Context-Aware Mixture of Experts (CAMoE) network, which aligns video features with various textual aspects. The CAMoE network leverages a mixture of expert models to effectively capture the interactions between different textual aspects and video representations, leading to improved retrieval accuracy.
Wang et al. \cite{DBLP:conf/cvpr/WangZ021} presented the T2VLAD method, which incorporates global-local alignment for video retrieval. T2VLAD utilizes a combination of global and local features to better capture the spatial and temporal information in videos. By aligning these features with textual queries, T2VLAD enhances retrieval precision and robustness.
Liu et al. \cite{DBLP:conf/iccv/Liu0QCDW21} proposed a Hierarchical Transformer, which performs cross-modal hierarchical matching. It employs a hierarchical structure that captures semantic and temporal dependencies between video frames and textual queries. By considering the hierarchical context, the model improves the alignment accuracy and discriminability between video and text.
These fine-grained matching and alignment approaches have demonstrated their effectiveness in improving performance \cite{pei2023clipping,Liangke_Survey,li2023svitt}. By addressing the detailed correspondences between video and text, these models provide more accurate and contextually relevant video recommendations based on textual queries.
Nevertheless, the above existing methods ignore chunk-level matching between multi-modal chunks and thus restrict the performance gain.
There also exists other strands of work that introduce additional meta information such as video caption~\cite{wu2023cap4video}, video title~\cite{falcon2022feature}, and object features~\cite{liu2022featinter} to facilitate the text-video retrieval, and our proposed hypergraph network based matching strategy can be easily applied to these approaches.

\section{Conclusion}

This paper proposes a novel text-video retrieval model LEAN to capture chunk-level correlations by multi-modal hypergraph networks with variational inference.
Our approach involves constructing a multi-modal hypergraph for each text query and video, capturing n-ary interactions between different modalities. Moreover, to learn an optimal hypergraph structure, our model automatically learns the weights of the hyperedge and node in the hypergraph.
To further improve the model’s generalization ability, we feed the extracted features into a variational inference module, obtaining the variational representation under the Gaussian distribution. 
The incorporation of them allows our model to capture n-ary interactions among texts and videos and incorporate multiple type correlations.
% This allows our model to handle situations where there is uncertainty in the relationships between textual and visual modalities, resulting in more accurate text-video retrieval predictions.
The empirical experiments demonstrate that our method incorporates chunk-level correlations and achieves state-of-the-art performance. 

% However, the operator of attribute generation will introduce noise data. 
% In future work, we will study how to avoid the influence of noise data on the MMEA task.


\begin{acks}
We thank the anonymous reviewers for their insightful comments and suggestions. 
XXX is the corresponding author.
The authors of this paper were supported by the NSFC through grant No.XXX. 

\end{acks}

\bibliographystyle{ACM-Reference-Format}
\bibliography{sample-base}


\begin{thebibliography}{38}

%%% the Web address.

\ifx \showCODEN    \undefined \def \showCODEN     #1{\unskip}     \fi
\ifx \showDOI      \undefined \def \showDOI       #1{#1}\fi
\ifx \showISBNx    \undefined \def \showISBNx     #1{\unskip}     \fi
\ifx \showISBNxiii \undefined \def \showISBNxiii  #1{\unskip}     \fi
\ifx \showISSN     \undefined \def \showISSN      #1{\unskip}     \fi
\ifx \showLCCN     \undefined \def \showLCCN      #1{\unskip}     \fi
\ifx \shownote     \undefined \def \shownote      #1{#1}          \fi
\ifx \showarticletitle \undefined \def \showarticletitle #1{#1}   \fi
\ifx \showURL      \undefined \def \showURL       {\relax}        \fi
% invisible to TeX
\providecommand\bibfield[2]{#2}
\providecommand\bibinfo[2]{#2}
\providecommand\natexlab[1]{#1}
\providecommand\showeprint[2][]{arXiv:#2}

\bibitem[Bain et~al\mbox{.}(2021)]%
        {DBLP:conf/iccv/BainNVZ21}
\bibfield{author}{\bibinfo{person}{Max Bain}, \bibinfo{person}{Arsha Nagrani},
  \bibinfo{person}{G{\"{u}}l Varol}, {and} \bibinfo{person}{Andrew Zisserman}.}
  \bibinfo{year}{2021}\natexlab{}.
\newblock \showarticletitle{Frozen in Time: {A} Joint Video and Image Encoder
  for End-to-End Retrieval}. In \bibinfo{booktitle}{\emph{2021 {IEEE/CVF}
  International Conference on Computer Vision, {ICCV} 2021, Montreal, QC,
  Canada, October 10-17, 2021}}. \bibinfo{publisher}{{IEEE}},
  \bibinfo{pages}{1708--1718}.
\newblock
\urldef\tempurl%
\url{https://doi.org/10.1109/ICCV48922.2021.00175}
\showDOI{\tempurl}


\bibitem[Chen and Dolan(2011)]%
        {chen2011collecting}
\bibfield{author}{\bibinfo{person}{David Chen} {and} \bibinfo{person}{William~B
  Dolan}.} \bibinfo{year}{2011}\natexlab{}.
\newblock \showarticletitle{Collecting highly parallel data for paraphrase
  evaluation}. In \bibinfo{booktitle}{\emph{Proceedings of the 49th annual
  meeting of the association for computational linguistics: human language
  technologies}}. \bibinfo{pages}{190--200}.
\newblock


\bibitem[Cheng et~al\mbox{.}(2021)]%
        {DBLP:journals/corr/abs-2109-04290}
\bibfield{author}{\bibinfo{person}{Xing Cheng}, \bibinfo{person}{Hezheng Lin},
  \bibinfo{person}{Xiangyu Wu}, \bibinfo{person}{Fan Yang}, {and}
  \bibinfo{person}{Dong Shen}.} \bibinfo{year}{2021}\natexlab{}.
\newblock \showarticletitle{Improving Video-Text Retrieval by Multi-Stream
  Corpus Alignment and Dual Softmax Loss}.
\newblock \bibinfo{journal}{\emph{CoRR}}  \bibinfo{volume}{abs/2109.04290}
  (\bibinfo{year}{2021}).
\newblock
\showeprint[arXiv]{2109.04290}


\bibitem[Choi and Rasmussen(2002)]%
        {choi2002users}
\bibfield{author}{\bibinfo{person}{Youngok Choi} {and} \bibinfo{person}{Edie~M
  Rasmussen}.} \bibinfo{year}{2002}\natexlab{}.
\newblock \showarticletitle{Users' relevance criteria in image retrieval in
  American history}.
\newblock \bibinfo{journal}{\emph{Information processing \& management}}
  \bibinfo{volume}{38}, \bibinfo{number}{5} (\bibinfo{year}{2002}),
  \bibinfo{pages}{695--726}.
\newblock


\bibitem[Croitoru et~al\mbox{.}(2021)]%
        {croitoru2021teachtext}
\bibfield{author}{\bibinfo{person}{Ioana Croitoru},
  \bibinfo{person}{Simion-Vlad Bogolin}, \bibinfo{person}{Marius Leordeanu},
  \bibinfo{person}{Hailin Jin}, \bibinfo{person}{Andrew Zisserman},
  \bibinfo{person}{Samuel Albanie}, {and} \bibinfo{person}{Yang Liu}.}
  \bibinfo{year}{2021}\natexlab{}.
\newblock \showarticletitle{Teachtext: Crossmodal generalized distillation for
  text-video retrieval}. In \bibinfo{booktitle}{\emph{Proceedings of the
  IEEE/CVF International Conference on Computer Vision}}.
  \bibinfo{pages}{11583--11593}.
\newblock


\bibitem[Devlin et~al\mbox{.}(2019)]%
        {DBLP:conf/naacl/DevlinCLT19}
\bibfield{author}{\bibinfo{person}{Jacob Devlin}, \bibinfo{person}{Ming{-}Wei
  Chang}, \bibinfo{person}{Kenton Lee}, {and} \bibinfo{person}{Kristina
  Toutanova}.} \bibinfo{year}{2019}\natexlab{}.
\newblock \showarticletitle{{BERT:} Pre-training of Deep Bidirectional
  Transformers for Language Understanding}. In
  \bibinfo{booktitle}{\emph{Proceedings of the 2019 Conference of the North
  American Chapter of the Association for Computational Linguistics: Human
  Language Technologies, {NAACL-HLT} 2019, Minneapolis, MN, USA, June 2-7,
  2019, Volume 1 (Long and Short Papers)}}. \bibinfo{pages}{4171--4186}.
\newblock


\bibitem[Dong et~al\mbox{.}(2022)]%
        {dong2022reading}
\bibfield{author}{\bibinfo{person}{Jianfeng Dong}, \bibinfo{person}{Yabing
  Wang}, \bibinfo{person}{Xianke Chen}, \bibinfo{person}{Xiaoye Qu},
  \bibinfo{person}{Xirong Li}, \bibinfo{person}{Yuan He}, {and}
  \bibinfo{person}{Xun Wang}.} \bibinfo{year}{2022}\natexlab{}.
\newblock \showarticletitle{Reading-strategy inspired visual representation
  learning for text-to-video retrieval}.
\newblock \bibinfo{journal}{\emph{IEEE transactions on circuits and systems for
  video technology}} \bibinfo{volume}{32}, \bibinfo{number}{8}
  (\bibinfo{year}{2022}), \bibinfo{pages}{5680--5694}.
\newblock


\bibitem[Falcon et~al\mbox{.}(2022)]%
        {falcon2022feature}
\bibfield{author}{\bibinfo{person}{Alex Falcon}, \bibinfo{person}{Giuseppe
  Serra}, {and} \bibinfo{person}{Oswald Lanz}.}
  \bibinfo{year}{2022}\natexlab{}.
\newblock \showarticletitle{A feature-space multimodal data augmentation
  technique for text-video retrieval}. In \bibinfo{booktitle}{\emph{Proceedings
  of the 30th ACM International Conference on Multimedia}}.
  \bibinfo{pages}{4385--4394}.
\newblock


\bibitem[Fang et~al\mbox{.}(2021)]%
        {DBLP:journals/corr/abs-2106-11097}
\bibfield{author}{\bibinfo{person}{Han Fang}, \bibinfo{person}{Pengfei Xiong},
  \bibinfo{person}{Luhui Xu}, {and} \bibinfo{person}{Yu Chen}.}
  \bibinfo{year}{2021}\natexlab{}.
\newblock \showarticletitle{CLIP2Video: Mastering Video-Text Retrieval via
  Image {CLIP}}.
\newblock \bibinfo{journal}{\emph{CoRR}}  \bibinfo{volume}{abs/2106.11097}
  (\bibinfo{year}{2021}).
\newblock
\showeprint[arXiv]{2106.11097}


\bibitem[Feng et~al\mbox{.}(2019)]%
        {feng2019hypergraph}
\bibfield{author}{\bibinfo{person}{Yifan Feng}, \bibinfo{person}{Haoxuan You},
  \bibinfo{person}{Zizhao Zhang}, \bibinfo{person}{Rongrong Ji}, {and}
  \bibinfo{person}{Yue Gao}.} \bibinfo{year}{2019}\natexlab{}.
\newblock \showarticletitle{Hypergraph neural networks}. In
  \bibinfo{booktitle}{\emph{Proceedings of the AAAI conference on artificial
  intelligence}}, Vol.~\bibinfo{volume}{33}. \bibinfo{pages}{3558--3565}.
\newblock


\bibitem[Gabeur et~al\mbox{.}(2020)]%
        {DBLP:conf/eccv/Gabeur0AS20}
\bibfield{author}{\bibinfo{person}{Valentin Gabeur}, \bibinfo{person}{Chen
  Sun}, \bibinfo{person}{Karteek Alahari}, {and} \bibinfo{person}{Cordelia
  Schmid}.} \bibinfo{year}{2020}\natexlab{}.
\newblock \showarticletitle{Multi-modal Transformer for Video Retrieval}. In
  \bibinfo{booktitle}{\emph{Computer Vision - {ECCV} 2020 - 16th European
  Conference, Glasgow, UK, August 23-28, 2020, Proceedings, Part {IV}}}
  \emph{(\bibinfo{series}{Lecture Notes in Computer Science},
  Vol.~\bibinfo{volume}{12349})}. \bibinfo{publisher}{Springer},
  \bibinfo{pages}{214--229}.
\newblock
\urldef\tempurl%
\url{https://doi.org/10.1007/978-3-030-58548-8\_13}
\showDOI{\tempurl}


\bibitem[Guo et~al\mbox{.}(2020)]%
        {guo2020deep}
\bibfield{author}{\bibinfo{person}{Jiafeng Guo}, \bibinfo{person}{Yixing Fan},
  \bibinfo{person}{Liang Pang}, \bibinfo{person}{Liu Yang},
  \bibinfo{person}{Qingyao Ai}, \bibinfo{person}{Hamed Zamani},
  \bibinfo{person}{Chen Wu}, \bibinfo{person}{W~Bruce Croft}, {and}
  \bibinfo{person}{Xueqi Cheng}.} \bibinfo{year}{2020}\natexlab{}.
\newblock \showarticletitle{A deep look into neural ranking models for
  information retrieval}.
\newblock \bibinfo{journal}{\emph{Information Processing \& Management}}
  \bibinfo{volume}{57}, \bibinfo{number}{6} (\bibinfo{year}{2020}),
  \bibinfo{pages}{102067}.
\newblock


\bibitem[Guo et~al\mbox{.}(2021)]%
        {guo2021ssan}
\bibfield{author}{\bibinfo{person}{Xudong Guo}, \bibinfo{person}{Xun Guo},
  {and} \bibinfo{person}{Yan Lu}.} \bibinfo{year}{2021}\natexlab{}.
\newblock \showarticletitle{Ssan: Separable self-attention network for video
  representation learning}. In \bibinfo{booktitle}{\emph{Proceedings of the
  IEEE/CVF conference on computer vision and pattern recognition}}.
  \bibinfo{pages}{12618--12627}.
\newblock


\bibitem[Han et~al\mbox{.}(2021)]%
        {han2021visual}
\bibfield{author}{\bibinfo{person}{Ning Han}, \bibinfo{person}{Jingjing Chen},
  \bibinfo{person}{Guangyi Xiao}, \bibinfo{person}{Yawen Zeng},
  \bibinfo{person}{Chuhao Shi}, {and} \bibinfo{person}{Hao Chen}.}
  \bibinfo{year}{2021}\natexlab{}.
\newblock \showarticletitle{Visual spatio-temporal relation-enhanced network
  for cross-modal text-video retrieval}.
\newblock \bibinfo{journal}{\emph{arXiv preprint arXiv:2110.15609}}
  (\bibinfo{year}{2021}).
\newblock


\bibitem[Jin et~al\mbox{.}(2023)]%
        {DBLP:journals/corr/abs-2305-12218}
\bibfield{author}{\bibinfo{person}{Peng Jin}, \bibinfo{person}{Hao Li},
  \bibinfo{person}{Zesen Cheng}, \bibinfo{person}{Jinfa Huang},
  \bibinfo{person}{Zhennan Wang}, \bibinfo{person}{Li Yuan},
  \bibinfo{person}{Chang Liu}, {and} \bibinfo{person}{Jie Chen}.}
  \bibinfo{year}{2023}\natexlab{}.
\newblock \showarticletitle{Text-Video Retrieval with Disentangled
  Conceptualization and Set-to-Set Alignment}.
\newblock \bibinfo{journal}{\emph{CoRR}}  \bibinfo{volume}{abs/2305.12218}
  (\bibinfo{year}{2023}).
\newblock
\urldef\tempurl%
\url{https://doi.org/10.48550/arXiv.2305.12218}
\showDOI{\tempurl}
\showeprint[arXiv]{2305.12218}


\bibitem[Lei et~al\mbox{.}(2021)]%
        {lei2021less}
\bibfield{author}{\bibinfo{person}{Jie Lei}, \bibinfo{person}{Linjie Li},
  \bibinfo{person}{Luowei Zhou}, \bibinfo{person}{Zhe Gan},
  \bibinfo{person}{Tamara~L Berg}, \bibinfo{person}{Mohit Bansal}, {and}
  \bibinfo{person}{Jingjing Liu}.} \bibinfo{year}{2021}\natexlab{}.
\newblock \showarticletitle{Less is more: Clipbert for video-and-language
  learning via sparse sampling}. In \bibinfo{booktitle}{\emph{Proceedings of
  the IEEE/CVF conference on computer vision and pattern recognition}}.
  \bibinfo{pages}{7331--7341}.
\newblock


\bibitem[Li et~al\mbox{.}(2020)]%
        {DBLP:conf/emnlp/LiCCGYL20}
\bibfield{author}{\bibinfo{person}{Linjie Li}, \bibinfo{person}{Yen{-}Chun
  Chen}, \bibinfo{person}{Yu Cheng}, \bibinfo{person}{Zhe Gan},
  \bibinfo{person}{Licheng Yu}, {and} \bibinfo{person}{Jingjing Liu}.}
  \bibinfo{year}{2020}\natexlab{}.
\newblock \showarticletitle{{HERO:} Hierarchical Encoder for Video+Language
  Omni-representation Pre-training}. In \bibinfo{booktitle}{\emph{Proceedings
  of the 2020 Conference on Empirical Methods in Natural Language Processing,
  {EMNLP} 2020, Online, November 16-20, 2020}}. \bibinfo{publisher}{Association
  for Computational Linguistics}, \bibinfo{pages}{2046--2065}.
\newblock
\urldef\tempurl%
\url{https://doi.org/10.18653/v1/2020.emnlp-main.161}
\showDOI{\tempurl}


\bibitem[Li et~al\mbox{.}(2023)]%
        {li2023svitt}
\bibfield{author}{\bibinfo{person}{Yi Li}, \bibinfo{person}{Kyle Min},
  \bibinfo{person}{Subarna Tripathi}, {and} \bibinfo{person}{Nuno
  Vasconcelos}.} \bibinfo{year}{2023}\natexlab{}.
\newblock \showarticletitle{SViTT: Temporal Learning of Sparse Video-Text
  Transformers}. In \bibinfo{booktitle}{\emph{Proceedings of the IEEE/CVF
  Conference on Computer Vision and Pattern Recognition}}.
  \bibinfo{pages}{18919--18929}.
\newblock


\bibitem[Liang et~al\mbox{.}(2022)]%
        {Liangke_Survey}
\bibfield{author}{\bibinfo{person}{Ke Liang}, \bibinfo{person}{Lingyuan Meng},
  \bibinfo{person}{Meng Liu}, \bibinfo{person}{Yue Liu},
  \bibinfo{person}{Wenxuan Tu}, \bibinfo{person}{Siwei Wang},
  \bibinfo{person}{Sihang Zhou}, \bibinfo{person}{Xinwang Liu}, {and}
  \bibinfo{person}{Fuchun Sun}.} \bibinfo{year}{2022}\natexlab{}.
\newblock \showarticletitle{Reasoning over different types of knowledge graphs:
  Static, temporal and multi-modal}.
\newblock \bibinfo{journal}{\emph{arXiv preprint arXiv:2212.05767}}
  (\bibinfo{year}{2022}).
\newblock


\bibitem[Liu et~al\mbox{.}(2022b)]%
        {liu2022featinter}
\bibfield{author}{\bibinfo{person}{Baolong Liu}, \bibinfo{person}{Qi Zheng},
  \bibinfo{person}{Yabing Wang}, \bibinfo{person}{Minsong Zhang},
  \bibinfo{person}{Jianfeng Dong}, {and} \bibinfo{person}{Xun Wang}.}
  \bibinfo{year}{2022}\natexlab{b}.
\newblock \showarticletitle{FeatInter: exploring fine-grained object features
  for video-text retrieval}.
\newblock \bibinfo{journal}{\emph{Neurocomputing}}  \bibinfo{volume}{496}
  (\bibinfo{year}{2022}), \bibinfo{pages}{178--191}.
\newblock


\bibitem[Liu et~al\mbox{.}(2021)]%
        {DBLP:conf/iccv/Liu0QCDW21}
\bibfield{author}{\bibinfo{person}{Song Liu}, \bibinfo{person}{Haoqi Fan},
  \bibinfo{person}{Shengsheng Qian}, \bibinfo{person}{Yiru Chen},
  \bibinfo{person}{Wenkui Ding}, {and} \bibinfo{person}{Zhongyuan Wang}.}
  \bibinfo{year}{2021}\natexlab{}.
\newblock \showarticletitle{HiT: Hierarchical Transformer with Momentum
  Contrast for Video-Text Retrieval}. In \bibinfo{booktitle}{\emph{2021
  {IEEE/CVF} International Conference on Computer Vision, {ICCV} 2021,
  Montreal, QC, Canada, October 10-17, 2021}}. \bibinfo{publisher}{{IEEE}},
  \bibinfo{pages}{11895--11905}.
\newblock
\urldef\tempurl%
\url{https://doi.org/10.1109/ICCV48922.2021.01170}
\showDOI{\tempurl}


\bibitem[Liu et~al\mbox{.}(2019)]%
        {DBLP:conf/bmvc/LiuANZ19}
\bibfield{author}{\bibinfo{person}{Yang Liu}, \bibinfo{person}{Samuel Albanie},
  \bibinfo{person}{Arsha Nagrani}, {and} \bibinfo{person}{Andrew Zisserman}.}
  \bibinfo{year}{2019}\natexlab{}.
\newblock \showarticletitle{Use What You Have: Video retrieval using
  representations from collaborative experts}. In
  \bibinfo{booktitle}{\emph{30th British Machine Vision Conference 2019, {BMVC}
  2019, Cardiff, UK, September 9-12, 2019}}. \bibinfo{publisher}{{BMVA} Press},
  \bibinfo{pages}{279}.
\newblock


\bibitem[Liu et~al\mbox{.}(2022a)]%
        {DBLP:conf/sigir/LiuCHCWPW22}
\bibfield{author}{\bibinfo{person}{Yu Liu}, \bibinfo{person}{Huai Chen},
  \bibinfo{person}{Lianghua Huang}, \bibinfo{person}{Di Chen},
  \bibinfo{person}{Bin Wang}, \bibinfo{person}{Pan Pan}, {and}
  \bibinfo{person}{Lisheng Wang}.} \bibinfo{year}{2022}\natexlab{a}.
\newblock \showarticletitle{Animating Images to Transfer {CLIP} for Video-Text
  Retrieval}. In \bibinfo{booktitle}{\emph{{SIGIR} '22: The 45th International
  {ACM} {SIGIR} Conference on Research and Development in Information
  Retrieval, Madrid, Spain, July 11 - 15, 2022}}. \bibinfo{publisher}{{ACM}},
  \bibinfo{pages}{1906--1911}.
\newblock
\urldef\tempurl%
\url{https://doi.org/10.1145/3477495.3531776}
\showDOI{\tempurl}


\bibitem[Luo et~al\mbox{.}(2022)]%
        {DBLP:journals/ijon/LuoJZCLDL22}
\bibfield{author}{\bibinfo{person}{Huaishao Luo}, \bibinfo{person}{Lei Ji},
  \bibinfo{person}{Ming Zhong}, \bibinfo{person}{Yang Chen},
  \bibinfo{person}{Wen Lei}, \bibinfo{person}{Nan Duan}, {and}
  \bibinfo{person}{Tianrui Li}.} \bibinfo{year}{2022}\natexlab{}.
\newblock \showarticletitle{CLIP4Clip: An empirical study of {CLIP} for end to
  end video clip retrieval and captioning}.
\newblock \bibinfo{journal}{\emph{Neurocomputing}}  \bibinfo{volume}{508}
  (\bibinfo{year}{2022}), \bibinfo{pages}{293--304}.
\newblock
\urldef\tempurl%
\url{https://doi.org/10.1016/j.neucom.2022.07.028}
\showDOI{\tempurl}


\bibitem[Ma et~al\mbox{.}(2022)]%
        {DBLP:conf/mm/MaXSYZJ22}
\bibfield{author}{\bibinfo{person}{Yiwei Ma}, \bibinfo{person}{Guohai Xu},
  \bibinfo{person}{Xiaoshuai Sun}, \bibinfo{person}{Ming Yan},
  \bibinfo{person}{Ji Zhang}, {and} \bibinfo{person}{Rongrong Ji}.}
  \bibinfo{year}{2022}\natexlab{}.
\newblock \showarticletitle{{X-CLIP:} End-to-End Multi-grained Contrastive
  Learning for Video-Text Retrieval}. In \bibinfo{booktitle}{\emph{{MM} '22:
  The 30th {ACM} International Conference on Multimedia, Lisboa, Portugal,
  October 10 - 14, 2022}}. \bibinfo{publisher}{{ACM}},
  \bibinfo{pages}{638--647}.
\newblock
\urldef\tempurl%
\url{https://doi.org/10.1145/3503161.3547910}
\showDOI{\tempurl}


\bibitem[Patrick et~al\mbox{.}(2021)]%
        {DBLP:conf/iclr/Patrick0AMHHV21}
\bibfield{author}{\bibinfo{person}{Mandela Patrick}, \bibinfo{person}{Po{-}Yao
  Huang}, \bibinfo{person}{Yuki~Markus Asano}, \bibinfo{person}{Florian Metze},
  \bibinfo{person}{Alexander~G. Hauptmann}, \bibinfo{person}{Jo{\~{a}}o~F.
  Henriques}, {and} \bibinfo{person}{Andrea Vedaldi}.}
  \bibinfo{year}{2021}\natexlab{}.
\newblock \showarticletitle{Support-set bottlenecks for video-text
  representation learning}. In \bibinfo{booktitle}{\emph{9th International
  Conference on Learning Representations, {ICLR} 2021, Virtual Event, Austria,
  May 3-7, 2021}}. \bibinfo{publisher}{OpenReview.net}.
\newblock


\bibitem[Pei et~al\mbox{.}(2023)]%
        {pei2023clipping}
\bibfield{author}{\bibinfo{person}{Renjing Pei}, \bibinfo{person}{Jianzhuang
  Liu}, \bibinfo{person}{Weimian Li}, \bibinfo{person}{Bin Shao},
  \bibinfo{person}{Songcen Xu}, \bibinfo{person}{Peng Dai},
  \bibinfo{person}{Juwei Lu}, {and} \bibinfo{person}{Youliang Yan}.}
  \bibinfo{year}{2023}\natexlab{}.
\newblock \showarticletitle{CLIPPING: Distilling CLIP-Based Models with a
  Student Base for Video-Language Retrieval}. In
  \bibinfo{booktitle}{\emph{Proceedings of the IEEE/CVF Conference on Computer
  Vision and Pattern Recognition}}. \bibinfo{pages}{18983--18992}.
\newblock


\bibitem[Radford et~al\mbox{.}(2021)]%
        {DBLP:conf/icml/RadfordKHRGASAM21}
\bibfield{author}{\bibinfo{person}{Alec Radford}, \bibinfo{person}{Jong~Wook
  Kim}, \bibinfo{person}{Chris Hallacy}, \bibinfo{person}{Aditya Ramesh},
  \bibinfo{person}{Gabriel Goh}, \bibinfo{person}{Sandhini Agarwal},
  \bibinfo{person}{Girish Sastry}, \bibinfo{person}{Amanda Askell},
  \bibinfo{person}{Pamela Mishkin}, \bibinfo{person}{Jack Clark},
  \bibinfo{person}{Gretchen Krueger}, {and} \bibinfo{person}{Ilya Sutskever}.}
  \bibinfo{year}{2021}\natexlab{}.
\newblock \showarticletitle{Learning Transferable Visual Models From Natural
  Language Supervision}. In \bibinfo{booktitle}{\emph{Proceedings of the 38th
  International Conference on Machine Learning, {ICML} 2021, 18-24 July 2021,
  Virtual Event}} \emph{(\bibinfo{series}{Proceedings of Machine Learning
  Research}, Vol.~\bibinfo{volume}{139})}. \bibinfo{publisher}{{PMLR}},
  \bibinfo{pages}{8748--8763}.
\newblock


\bibitem[Soares et~al\mbox{.}(2019)]%
        {DBLP:conf/acl/SoaresFLK19}
\bibfield{author}{\bibinfo{person}{Livio~Baldini Soares},
  \bibinfo{person}{Nicholas FitzGerald}, \bibinfo{person}{Jeffrey Ling}, {and}
  \bibinfo{person}{Tom Kwiatkowski}.} \bibinfo{year}{2019}\natexlab{}.
\newblock \showarticletitle{Matching the Blanks: Distributional Similarity for
  Relation Learning}. In \bibinfo{booktitle}{\emph{Proceedings of the 57th
  Conference of the Association for Computational Linguistics, {ACL} 2019,
  Florence, Italy, July 28- August 2, 2019, Volume 1: Long Papers}}.
  \bibinfo{publisher}{Association for Computational Linguistics},
  \bibinfo{pages}{2895--2905}.
\newblock
\urldef\tempurl%
\url{https://doi.org/10.18653/v1/p19-1279}
\showDOI{\tempurl}


\bibitem[Wang et~al\mbox{.}(2022)]%
        {wang2022hybrid}
\bibfield{author}{\bibinfo{person}{Jinpeng Wang}, \bibinfo{person}{Bin Chen},
  \bibinfo{person}{Dongliang Liao}, \bibinfo{person}{Ziyun Zeng},
  \bibinfo{person}{Gongfu Li}, \bibinfo{person}{Shu-Tao Xia}, {and}
  \bibinfo{person}{Jin Xu}.} \bibinfo{year}{2022}\natexlab{}.
\newblock \showarticletitle{Hybrid contrastive quantization for efficient
  cross-view video retrieval}. In \bibinfo{booktitle}{\emph{Proceedings of the
  ACM Web Conference 2022}}. \bibinfo{pages}{3020--3030}.
\newblock


\bibitem[Wang et~al\mbox{.}(2021)]%
        {DBLP:conf/cvpr/WangZ021}
\bibfield{author}{\bibinfo{person}{Xiaohan Wang}, \bibinfo{person}{Linchao
  Zhu}, {and} \bibinfo{person}{Yi Yang}.} \bibinfo{year}{2021}\natexlab{}.
\newblock \showarticletitle{{T2VLAD:} Global-Local Sequence Alignment for
  Text-Video Retrieval}. In \bibinfo{booktitle}{\emph{{IEEE} Conference on
  Computer Vision and Pattern Recognition, {CVPR} 2021, virtual, June 19-25,
  2021}}. \bibinfo{publisher}{Computer Vision Foundation / {IEEE}},
  \bibinfo{pages}{5079--5088}.
\newblock
\urldef\tempurl%
\url{https://doi.org/10.1109/CVPR46437.2021.00504}
\showDOI{\tempurl}


\bibitem[Wu et~al\mbox{.}(2021)]%
        {wu2021hanet}
\bibfield{author}{\bibinfo{person}{Peng Wu}, \bibinfo{person}{Xiangteng He},
  \bibinfo{person}{Mingqian Tang}, \bibinfo{person}{Yiliang Lv}, {and}
  \bibinfo{person}{Jing Liu}.} \bibinfo{year}{2021}\natexlab{}.
\newblock \showarticletitle{Hanet: Hierarchical alignment networks for
  video-text retrieval}. In \bibinfo{booktitle}{\emph{Proceedings of the 29th
  ACM international conference on Multimedia}}. \bibinfo{pages}{3518--3527}.
\newblock


\bibitem[Wu et~al\mbox{.}(2023)]%
        {wu2023cap4video}
\bibfield{author}{\bibinfo{person}{Wenhao Wu}, \bibinfo{person}{Haipeng Luo},
  \bibinfo{person}{Bo Fang}, \bibinfo{person}{Jingdong Wang}, {and}
  \bibinfo{person}{Wanli Ouyang}.} \bibinfo{year}{2023}\natexlab{}.
\newblock \showarticletitle{Cap4Video: What Can Auxiliary Captions Do for
  Text-Video Retrieval?}. In \bibinfo{booktitle}{\emph{Proceedings of the
  IEEE/CVF Conference on Computer Vision and Pattern Recognition}}.
  \bibinfo{pages}{10704--10713}.
\newblock


\bibitem[Xu et~al\mbox{.}(2016)]%
        {xu2016msr}
\bibfield{author}{\bibinfo{person}{Jun Xu}, \bibinfo{person}{Tao Mei},
  \bibinfo{person}{Ting Yao}, {and} \bibinfo{person}{Yong Rui}.}
  \bibinfo{year}{2016}\natexlab{}.
\newblock \showarticletitle{Msr-vtt: A large video description dataset for
  bridging video and language}. In \bibinfo{booktitle}{\emph{Proceedings of the
  IEEE conference on computer vision and pattern recognition}}.
  \bibinfo{pages}{5288--5296}.
\newblock


\bibitem[Yakovlev et~al\mbox{.}(2023)]%
        {DBLP:conf/sigir/YakovlevPAPBNP23}
\bibfield{author}{\bibinfo{person}{Konstantin Yakovlev},
  \bibinfo{person}{Gregory Polyakov}, \bibinfo{person}{Ilseyar Alimova},
  \bibinfo{person}{Alexander Podolskiy}, \bibinfo{person}{Andrey Bout},
  \bibinfo{person}{Sergey Nikolenko}, {and} \bibinfo{person}{Irina
  Piontkovskaya}.} \bibinfo{year}{2023}\natexlab{}.
\newblock \showarticletitle{Sinkhorn Transformations for Single-Query
  Postprocessing in Text-Video Retrieval}. In
  \bibinfo{booktitle}{\emph{Proceedings of the 46th International {ACM} {SIGIR}
  Conference on Research and Development in Information Retrieval, {SIGIR}
  2023, Taipei, Taiwan, July 23-27, 2023}}. \bibinfo{publisher}{{ACM}},
  \bibinfo{pages}{2394--2398}.
\newblock
\urldef\tempurl%
\url{https://doi.org/10.1145/3539618.3592064}
\showDOI{\tempurl}


\bibitem[Yang et~al\mbox{.}(2021)]%
        {DBLP:conf/iccv/YangBG21}
\bibfield{author}{\bibinfo{person}{Jianwei Yang}, \bibinfo{person}{Yonatan
  Bisk}, {and} \bibinfo{person}{Jianfeng Gao}.}
  \bibinfo{year}{2021}\natexlab{}.
\newblock \showarticletitle{TACo: Token-aware Cascade Contrastive Learning for
  Video-Text Alignment}. In \bibinfo{booktitle}{\emph{2021 {IEEE/CVF}
  International Conference on Computer Vision, {ICCV} 2021, Montreal, QC,
  Canada, October 10-17, 2021}}. \bibinfo{publisher}{{IEEE}},
  \bibinfo{pages}{11542--11552}.
\newblock
\urldef\tempurl%
\url{https://doi.org/10.1109/ICCV48922.2021.01136}
\showDOI{\tempurl}


\bibitem[Zhu et~al\mbox{.}(2023)]%
        {DBLP:journals/ijmir/ZhuJCGL23}
\bibfield{author}{\bibinfo{person}{Cunjuan Zhu}, \bibinfo{person}{Qi Jia},
  \bibinfo{person}{Wei Chen}, \bibinfo{person}{Yanming Guo}, {and}
  \bibinfo{person}{Yu Liu}.} \bibinfo{year}{2023}\natexlab{}.
\newblock \showarticletitle{Deep learning for video-text retrieval: a review}.
\newblock \bibinfo{journal}{\emph{Int. J. Multim. Inf. Retr.}}
  \bibinfo{volume}{12}, \bibinfo{number}{1} (\bibinfo{year}{2023}),
  \bibinfo{pages}{3}.
\newblock
\urldef\tempurl%
\url{https://doi.org/10.1007/s13735-023-00267-8}
\showDOI{\tempurl}


\bibitem[Zubair et~al\mbox{.}(2013)]%
        {DBLP:journals/dsp/ZubairYW13}
\bibfield{author}{\bibinfo{person}{Syed Zubair}, \bibinfo{person}{Fei Yan},
  {and} \bibinfo{person}{Wenwu Wang}.} \bibinfo{year}{2013}\natexlab{}.
\newblock \showarticletitle{Dictionary learning based sparse coefficients for
  audio classification with max and average pooling}.
\newblock \bibinfo{journal}{\emph{Digit. Signal Process.}}
  \bibinfo{volume}{23}, \bibinfo{number}{3} (\bibinfo{year}{2013}),
  \bibinfo{pages}{960--970}.
\newblock
\urldef\tempurl%
\url{https://doi.org/10.1016/j.dsp.2013.01.004}
\showDOI{\tempurl}


\end{thebibliography}

\end{document}